# Algorithmic encoding of protected characteristics in image-based models for disease detection

Ben Glocker, PhD* · Charles Jones, MEng · Mélanie Bernhardt, MSc · Stefan Winzeck, PhD

Department of Computing, Imperial College London, London, SW7 2AZ, United Kingdom
*Corresponding author: b.glocker@imperial.ac.uk

## Summary

### Background

It has been rightfully emphasized that the use of AI for clinical decision making could amplify health disparities. An algorithm may encode protected characteristics, and then use this information for making predictions due to undesirable correlations in the (historical) training data. It remains unclear how we can establish whether such information is actually used. Besides the scarcity of data from underserved populations, very little is known about how dataset biases manifest in predictive models and how this may result in disparate performance. This article aims to shed some light on these issues by exploring new methodology for subgroup analysis in image-based disease detection models.

### Methods

We utilize two publicly available chest X-ray datasets, CheXpert and MIMIC-CXR, to study performance disparities across race and biological sex in deep learning models. We explore test set resampling, transfer learning, multitask learning, and model inspection to assess the relationship between the encoding of protected characteristics and disease detection performance across subgroups.

### Findings

We confirm subgroup disparities in terms of shifted true and false positive rates which are partially removed after correcting for population and prevalence shifts in the test sets. We further find a previously used transfer learning method to be insufficient for establishing whether specific patient information is used for making predictions. The proposed combination of test-set resampling, multitask learning, and model inspection reveals valuable new insights about the way protected characteristics are encoded in the feature representations of deep neural networks.

### Interpretation

Subgroup analysis is key for identifying performance disparities of AI models, but statistical differences across subgroups need to be taken into account when analyzing potential biases in disease detection. The proposed methodology provides a comprehensive framework for subgroup analysis enabling further research into the underlying causes of disparities.

### Funding

European Research Council Horizon 2020, UK Research and Innovation.

# Introduction

Knowing what type of information is being used when machine learning (ML) models make predictions is of high relevance to all stakeholders, including clinicians, patients, developers, regulators, and policy makers.[1] Some of the best performing ML models, however, seem rather opaque regarding the mechanism by which these models map input data (e.g., medical scans) to output predictions (e.g., clinical diagnosis). During training, these advanced models, commonly based on deep neural networks, construct a rather complex feature representation by successively applying non-linear transformations to the raw input data. The final representations in the penultimate layer of the neural network are then processed by the last layer which acts as the prediction layer. The prediction layer learns to assign weights to the individual features in the penultimate layer which are then aggregated to generate an output prediction (e.g., a probability for the presence of disease). The basic components of a typical deep neural network model are illustrated in Figure 1.[2]

A key issue with these 'black box' models is that it is difficult to know what type of information is being used. The reason is that the generated feature representations encoded in the penultimate layer do not carry semantic information. So, we cannot easily check whether a particular type of information, say a patient's racial identity, is being used by the prediction layer, even if we know that such information is present in the original input data, either explicitly, e.g., in tabular data, or implicitly, e.g., in medical scans. An important point we make in this article is that even the presence of such information in the learned feature representation of the penultimate layer is not a sufficient indication that this information is being used by the prediction layer. We illustrate this with some concrete examples and a real world, clinical application of disease detection in chest X-ray.

## The unreasonable performance of deep learning

Despite the opaque behavior of deep neural networks, these models are now ubiquitous, and have become the state-of-the-art approach for most image-based prediction tasks.[3] An intriguing example for the 'power' of deep learning is the discovery that cardiovascular risk factors can be accurately predicted from retinal fundus images, including age, biological sex, smoking status, systolic blood pressure and major adverse cardiac events.[4] Additional biomarkers were discovered shortly after.[5] Another remarkable finding was that deep neural networks were capable of predicting patients' experienced pain from knee X-rays enabling an algorithmic approach for reducing unexplained racial disparities in pain.[6] Here, the technique of deep learning does not only help us to discover such previously unknown associations between 'raw' inputs and outputs but also to capture them in compact mathematical models that we can use for making accurate predictions on new data.[2]

In addition to being able to predict patient's age and biological sex,[7] a recent study further demonstrated that deep neural networks are also capable of recognizing a patient's racial identity from chest X-ray and other medical scans with astonishing accuracy.[8] The 'reading race' study is remarkable in multiple ways. Not only was it unknown that it is possible to recognize racial identity from these scans, it also appears to be a task that expert radiologists are not capable of doing (or at least not trained for). The exact mechanism and types of imaging features that are being used for making these predictions are yet to be uncovered. But more importantly, these results may have profound implications in the context discussed earlier, that there is a real risk that ML models may amplify health disparities.[9,10] Since it seems straightforward (given the very high accuracy) to extract features related to racial information from medical scans, any spurious correlations between race and clinical outcome present in the data could be picked up by a model that is trained for clinical diagnosis. Assuming that features predictive of race are easier to extract than features associated with pathology, the model is very likely to learn so called 'shortcuts', that would manifest an undesirable association in the model between the patient's race and the prediction of disease.[11] This emphasizes again the importance for being able to know what information is being used when a model makes predictions.

The 'reading race' study made some attempts at this, despite this not being the main focus of the work.[8] More concretely, the authors tried to establish whether the feature representations of deep neural network trained for disease detection, are useful for predicting race. To cite from the paper, the authors "hypothesized that if the [disease] model was able to identify a patient's race, this would suggest the model had implicitly learned to recognize racial information despite not being directly trained for that task." The study found this to be the case, using a specific type of test based on transfer learning that we will discuss below, and further concluded "these results suggest that even when race is poorly correlated with the outcome of interest, as is the case in pathology detection, deep learning models are likely to learn unintended cues related to race and incorporate these cues in their decision making."

This rather daunting conclusion is of particular concern in the light of recent studies that found performance disparities across racial subgroups.[12,13] The connection between the ability to predict patient characteristics from images and subgroup disparities in disease detection models demands further investigation as the implications are of high relevance for the safe and ethical use of AI.

## The inter-relationship of prediction tasks

One seemingly intuitive approach to investigate whether particular information is being used for making predictions is to check if the model (or more precisely, the learned feature representations) trained for a primary task of disease detection can be used for a secondary task for predicting patient characteristics. Assuming the secondary task can be performed reasonably well, one may conclude that the two tasks are closely related.

Here, we need to first clarify what we mean with "tasks are related". From a machine learning perspective, one may distinguish between three scenarios as illustrated in Figure 2 with the example of separating colors and shapes. In scenario A, the two tasks are unrelated both on the feature- and the output-level; In this case, we can solve each task independently using a different set of features and no information about the other task is relevant nor helpful. In scenario B, the tasks are related on a feature-level but not on an output-level; Here, the two tasks make use of the same feature representation, but apply different weights and aggregations for making the predictions. The information about one task, however, remains irrelevant for the other. In scenario C, the tasks are related both on a feature- and output-level, and it appears impossible to disentangle the tasks. Solving one task will, at least to some degree, also solve the other task. We can say that the information related to each task is correlated with the other task. In practice, there can be of course different degrees of correlation.

To establish whether certain information (say shape) is being used for solving the primary task (say color classification), we need to identify which is the relevant scenario. Indeed, we can see that in scenario A, the feature representation from the primary task (here just a single feature $x_1$) is not useful for solving the secondary task. We thus may safely conclude that shape has no role to play when classifying color and vice versa in this given example. In scenario C, we would find the opposite. The secondary task can be solved using the model learned for the primary task (and vice versa). We may conclude that there is no way of disentangling shape and color information, and a model trained for one task may use the information related to the other task. Scenario B is the most intriguing one. Here, we would find that the secondary task can be solved by using the features from the primary task. However, we need to learn a new set of feature weights specific to the secondary task, as the weights will be different from the ones learned for the primary task. So, while the information about one task is neither relevant nor helpful for the other task, we can still solve each task by using the features from the other task. This is an important observation which is relevant in the context of our real world application of disease detection. In cases where we have knowledge that the secondary task information (say a patient's race) should not be used for the primary task (say detection of disease), we need to avoid using models under scenario C. Models under scenario B, however, are potentially safe to use despite the fact that their feature representations could be predictive for the secondary task. In the following, we explore different methods including transfer learning, test set resampling, multitask learning, and model inspection to study the inner workings of disease detection models, drawing connections between their subgroup performance and the way protected characteristics are encoded.

## Methods

### Study population

We study the behavior of deep convolutional neural networks trained for detecting different conditions using two publicly available chest X-ray datasets, CheXpert and MIMIC-CXR.[14,15] The datasets contain detailed patient demographics including self-reported racial identity, biological sex, and age. The CheXpert sample contains a total of 42,884 patients with 127,118 chest X-ray scans divided into three sets for training (76,205), validation (12,673) and testing (38,240). The MIMIC-CXR sample contains 43,209 patients with 183,207 scans divided into training (110,280), validation (17,665), and testing (55,262). No scans from the same patient are used in different subsets. The validation sets are used for model selection, while the test sets are the hold out sets for measuring disease detection performance and assessing model behavior. Both CheXpert and MIMIC-CXR are highly imbalanced and skewed across subgroups. The large majority of scans are from patients identifying as White (78% and 77%), while scans from patients identifying as Asian (15% and 4%) and Black (7% and 19%) are underrepresented. Black patients in CheXpert are on average 5-8 years and in MIMIC-CXR 2-5 years younger than Asian and White patients. The proportion of females in CheXpert is 40%, 43% and 49% for White, Asian, and Black patients, and 43%, 44%, and 60% in MIMIC-CXR. The study samples and splits are identical to the ones used in the 'reading race' study.[8] A detailed breakdown of the population characteristics is provided in Table 1 with a visual summary provided in Figure S1 in the Supplementary Material.

### Disease detection models

The basis of our investigation is a real world, clinical application of image-based disease detection. We note that developing the disease detection models is not the primary concern of our study, nor do we claim any contribution in this respect. Here, we are studying models that have been used in previous works.[12,8,13] We train deep neural networks for detecting 14 different conditions annotated in the CheXpert and MIMIC-CXR datasets. Similar to previous work, we use a multi-label approach as patients may have multiple conditions. The presence of each condition is predicted via a dedicated one-dimensional output which is passed through a sigmoid function to obtain predictions between zero and one. The simultaneous detection of the individual conditions uses a common feature representation obtained from a shared neural network backbone. We focus our analysis on two tasks, one for detecting the presence of a specific pathology ('pleural effusion' label) and another aiming to rule out the presence of disease ('no finding' label). These two labels are mutually exclusive which makes them suitable for our model inspection, discussed later. The varying prevalence of these labels across subgroups and datasets makes them particularly interesting to study in the context of performance disparities under population and prevalence shifts (cf. Table 1 and Figure S1). The prevalence of 'no finding' is 8%, 9% and 10% for White, Asian, and Black patients in CheXpert, and 31%, 29%, and 38% in MIMIC-CXR. For 'pleural effusion', the prevalence is 41%, 42%, and 33% in CheXpert, and 27%, 27%, and 16% in MIMIC-CXR. The 'no finding' label was also the focus of a recent study which reported subgroup disparities assumed to be associated with underdiagnosis bias.[13] By using the same data and models, our performance analysis may shed further light on the reported issue of algorithmic bias.

**Test-set resampling for unbiased estimation of subgroup performance**

Previous studies have reported subgroup disparities in the form of shifted true and false positive rates in underserved populations, raising concerns that models may pick up bias from the training data which is then replicated at test-time.[12,13] A limitation in these studies, however, arises from their use of test data exhibiting the same biases as the training data (due to random splitting of the original datasets) which complicates the interpretation of the reported disparities.[16] Population and prevalence shifts across subgroups are known to cause disparities in predictive models.[17,18] In order to faithfully assess the behavior of a potentially biased model, one would require access to an unbiased test set which is difficult to obtain. For this reason, we explore the use of strategic resampling with replacement to construct balanced test sets that are representative of the population-of-interest.[19] Test set resampling allows us to correct for variations across subgroups such as racial imbalance, differences in age, and varying prevalence of disease. Controlling for specific characteristics when estimating subgroup performance and contrasting this with the performance found on the original test set may allow us to identify underlying causes of disparities. A visual summary of the population characteristics of the resampled test-sets is provided in Figure S2 in the Supplementary Material.

**The supervised prediction layer information test**

Besides subgroup performance, it is equally important to assess whether patient characteristics are encoded and then potentially used in a disease detection model. One approach to test this is based on transfer learning which has been used in the 'reading race' study to assess whether a disease detection model may have implicitly learned information about racial identity. This test can be easily implemented for deep neural networks by first training a neural network for a primary task, 'freezing' the network's parameters (i.e., the weights in what is often called the neural network 'backbone') and then replacing the prediction layer with a new one (cf. Figure 1). The new prediction layer is then trained specifically for the secondary task to learn a new set of weights assigned to the features in the penultimate layer. The features are generated by passing the input data through the frozen backbone from the primary task. We may then measure the accuracy of this new prediction layer on some test data. We might conclude that the two tasks are related and possibly even share information when the level of accuracy is reasonably high. In the following, we refer to this approach as the 'supervised prediction layer information test' or SPLIT.

We argue that SPLIT is insufficient to confirm whether a disease detection model may have implicitly learned to encode protected characteristics such as racial identity. Considering the example of shape and color classification. SPLIT can only tell us whether we are either in scenario A, in which case the accuracy obtained with SPLIT would have to be very low as the feature learned for one task is not useful for the other task (we call this a negative SPLIT result), or we are in one of the other two scenarios, B or C. However, SPLIT is unable to distinguish between those two. In fact, even the absolute value of the observed SPLIT accuracy is uninformative as we can easily confirm for the given example in scenario B (cf. Figure 2). Here, SPLIT would result in perfect classification of shape, despite shape being irrelevant for the classification of color. Obtaining a positive SPLIT result is a *necessary* condition for establishing that information is being used, but not a *sufficient* one. SPLIT is like a diagnostic test that has 100% sensitivity but unknown specificity.

To confirm these shortcomings, we applied SPLIT for both race and sex classification to different disease detection model backbones. Our main model uses the whole training set including all patients that identify as White, Asian, or Black. We trained two other disease detection models each using only a subgroup of patients to contrast our findings on the encoding of racial identity and biological sex. To minimize the effect of different amounts of training data on the model performance, we used the subgroups with the largest number of scans available, which are the groups of patients that identified as White, and male patients. The models trained on subgroups only are not exposed to the same variation in patient characteristics as the model trained on all data. In addition, we also considered two backbones that were neither trained for disease detection nor with any medical imaging data. One of them is based on random initialization of the network weights where the backbone then acts as a random projection of the input imaging data. This backbone is entirely untrained before applying SPLIT, and when combined with a prediction layer resembles a shallow model with limited capacity similar to a logistic regression. The motivation here is to provide baseline results for assessing the general difficulty of the tasks. The other non-medical backbone corresponds to a network trained for natural image classification using the ImageNet dataset.[20]

**Assessing task relationships via multitask learning**

While SPLIT cannot provide a definite answer whether specific information is being used or not, with a simple tweak to the prediction model, we can assess more explicitly the relationship of tasks. Recall that under scenario C where tasks are closely related both on the feature- and the output-level, the information from one task should be helpful for solving the other task. We can assess this by using the idea of multitask learning where we train a single model for simultaneously solving multiple prediction tasks.[21] This can be easily implemented with deep neural networks by using one shared neural network backbone and multiple prediction layers. In our setting of disease detection from chest X-rays, we can simply add two prediction layers to our model, one for predicting biological sex and one for racial identity, connected to the same penultimate layer of the backbone as the disease prediction layer. We then make explicit use of the labels for disease, sex, and race during training, to learn a feature representation that is shared across the three prediction tasks. If a patient's sex or racial identity is directly related to the prediction of disease (for example, due to unwanted correlations in the historical data), we may find that the task-specific features align in similar 'directions' in the feature space. Thus, inspecting the learned feature representation of a multitask model and comparing it to the feature space learned by a single task disease detection model may provide valuable insights about the inter-relationship of these prediction tasks.

**Unsupervised exploration of feature representations**

We employ a model inspection approach utilizing unsupervised machine learning techniques that allow us to directly explore what information is encoded, how it is distributed, and whether it aligns in the learned feature space with the primary task of disease detection. We recall that the prediction layer makes the ultimate decision about what information to use. The difficulty is that the feature representations are typically high-dimensional. In a DenseNet-121, the representations in the penultimate layer have 1024 dimensions.[22] To inspect the learned feature representations, we need to make use of dimensionality reduction techniques.

We use principal component analysis (PCA) to capture the main modes of variation within the feature representations. We then generate two-dimensional scatter plots for the first four PCA modes, and overlay different types of patient information. Additionally, we use t-distributed stochastic neighbor embedding (t-SNE), a popular algorithm for visualizing high-dimensional data, to capture the similarity between samples in the feature space.[23] We also plot the output predictions of the primary task prediction layer. In our case, the output of the disease detection model has 14 dimensions (one output for each of the 14 conditions). We may either apply dimensionality reduction on the 14-dimensional outputs, or focus on specific conditions of interest. Here, we focus on the two tasks of classifying 'no finding' and 'pleural effusion'. The two-dimensional logits (which are the unnormalized prediction scores for the two output classes) can then be directly visualized in a single scatter plot. Samples that are labeled neither 'no finding' nor 'pleural effusion' are labeled in the plots as 'other'. For each scatter plot of PCA, t-SNE and logit outputs, we also visualize the corresponding (marginal) distributions that one obtains when projecting the 2D data points against the axes of the scatter plots. We then visually check if any patterns emerge in these visualizations, and we use statistical tests to compare the marginal distributions in PCA and logit space for all relevant pairs of subgroups using the two-sample Kolmogorov-Smirnov test. Contrasting the encodings in the feature embeddings with the outputs of the prediction layer may allow us to assess whether particular information is used for making predictions.

**Statistical analysis**

The primary metrics for performance evaluation of the disease detection models include the area under the receiver operating characteristic curve (AUC), true positive rate (TPR), and false positive rate (FPR). TPR and FPR in subgroups are determined at a fixed decision threshold, which is optimized for each model to yield an FPR of 0·20 on the whole patient population. TPR is equal to sensitivity (and recall), while FPR is equal to 1 - specificity. We also report Youden's J statistic which is defined as J = sensitivity + specificity - 1, or simply J = TPR - FPR, providing a combined measure of classification performance. The relationship between TPR and FPR under different decision thresholds is illustrated in ROC curves. AUC and ROC curves allow the comparison of a model's classification ability independent of a specific decision threshold, while TPR and FPR allow the identification of threshold shifts causing subgroup disparities. SPLIT performance for race and sex classification is primarily measured with AUC. We also report TPR/FPR calculated for a decision threshold optimized for the highest Youden's J statistic. For the three-class race classification, we use a one-vs-rest approach for each racial group to measure classification performance. For all reported results, bootstrapping (stratified by targets) with 2,000 samples was used to calculate 95% confidence intervals.[24] Two-sample Kolmogorov-Smirnov tests are used in the unsupervised feature exploration to determine p-values for the null hypothesis that the marginal distributions for a pair of subgroups are identical in PCA and logit space.

**Role of the funding source**

**Ethical approval**

This research is exempt from ethical approval as the analysis is based on secondary data which is publicly available, and no permission is required to access the data.

# Results

## Disparities in disease detection performance

We observed good classification accuracy for a DenseNet-121 with an AUC of 0·87 (95% CI 0·87-0·88) and 0·85 (0·85-0·85) for 'no finding', and 0·87 (0·86-0·87) and 0·90 (0·89-0·90) for 'pleural effusion' on CheXpert and MIMIC-CXR, respectively, which is comparable to previous work.[12,13] A detailed breakdown across different patient subgroups and datasets is given in Tables 2 and 3 with corresponding ROC curves shown in Figures 3 and 4. Additional results for a ResNet-34 on the CheXpert dataset are provided in Table S1 in the Supplementary Material.

While AUC seems largely consistent across groups and ROC curves appear similar in shape, we observe some clear TPR/FPR shifts when using a fixed threshold for the entire population. On CheXpert, Black patients have an increased FPR of 3% for 'no finding' and a decreased FPR of 4% for 'pleural effusion' compared to the target value of 20%. On MIMIC-CXR, we observe an increased FPR of 5% for Black patients and a decreased FPR of 3% for Asian patients for 'no finding'. For 'pleural effusion', we also observe a decreased FPR of 5% for Black patients, and some shifts across biological sex. These findings seem to confirm that a decision threshold optimized over the whole patient population may not generalize and lead to disparate performance in underrepresented subgroups.[13] It is important to note, however, that most of the observed shifts do not imply that the models perform worse (or better) on some subgroups, but that they rather perform differently. A change in FPR usually comes with a corresponding change in TPR, reflected in largely consistent values for Youden's J statistic across subgroups. Nonetheless, such changes in model behavior would have important clinical implications. For example, a consistently increased FPR for 'no finding' would mean that in practice some patients are more likely to be underdiagnosed than others.[13]

A key question is whether the model may have implicitly learned to perform differently on different subgroups due to specific biases in the training data and the model's ability to recognize patient characteristics from the images. To assess the effect of the training characteristics on the subgroup disparities, we evaluated the performance of the CheXpert and MIMIC-CXR test sets each using the model trained on the other dataset. Because CheXpert and MIMIC-CXR vary substantially in the distribution of patient characteristics and prevalence of disease, we may expect to see differences compared to a model that was trained and tested on subsets from the same dataset. The results are given in Tables 4 and 5. We observe similar disparities across subgroups as before despite the different training characteristics of the CheXpert and MIMIC-CXR disease detection models. The shifts in TPR/FPR largely remain the same, suggesting that these disparities are potentially caused by the specific composition of the test sets, rather than by model bias.

To investigate the effect of the test set characteristics, we assessed performance of each disease detection model using strategic resampling to create race balanced test sets while controlling for age differences and disease prevalence in each racial group. The results are given in Tables 2 and 3 with the corresponding ROC curves shown in Figures 3 and 4. We find that using resampled test sets has a large effect on reducing TPR/FPR shifts for 'no finding' compared to the results from the original test sets. The FPR gets close to the target value of 0·2 across all subgroups and datasets while AUC and Youden's J statistic are largely preserved. For CheXpert, the previously observed disparities for 'no finding' seem to entirely disappear which may suggest that the shifts were primarily due to statistical differences in the test set subgroups rather than due to biases in the model. For the detection of 'pleural effusion', however, disparities remain. Black patients, in particular, have a reduced TPR/FPR across datasets and lower AUC and Youden's J statistics on CheXpert, both for the original and resampled test sets. This suggests that the models perform differently on this subgroup, and are in fact worse for detecting 'pleural effusion' in Black patients. The question is whether the models may implicitly use racial information for making predictions, which the following experiments aim to investigate.

**SPLIT performance**

The results for applying SPLIT to race classification on CheXpert are summarized in Figure 5 (with additional metrics provided in Table S2). We find that SPLIT is unable to detect clear differences between the models trained on all patients (with AUC for White 0·78 (95% CI 0·77-0·79), Asian 0·80 (0·80-0·81), and Black 0·78 (0·77-0·79)) versus the model trained only on White patients (with AUC for White 0·77 (0·76-0·77), Asian 0·79 (0·79-0·80), and Black 0·79 (0·78-0·80)), showing similar classification performance. We also observe a similarly high SPLIT response for the backbone trained on ImageNet (with AUC for White 0·77 (0·76-0·77), Asian 0·79 (0·79-0·80), and Black 0·78 (0·77-0·79)), which indicates that it is possible to successfully train a prediction layer to recognize race from chest X-ray using features learned for natural image classification. Here, the model trained on ImageNet, however, cannot have possibly learned to extract racial information as it has never seen any medical data, and yet, its features are useful for race classification when training a new prediction layer. In the case of ImageNet, we are likely seeing an example of scenario B discussed earlier, where the same features are useful for different tasks that are, however, unrelated on an output-level. We also observe higher than chance classification accuracy for the backbone with random weights (with AUC for White 0·70 (0·70-0·71), Asian 0·75 (0·74-0·76), and Black 0·67 (0·66-0·68)) which suggests that even random data transformations retain some signal from the raw images about racial identity. Figure 6 and Table S3 provide results for applying SPLIT to biological sex classification. Here, we replaced the backbone trained on White patients with a backbone trained on male patients. Similar to race classification, we obtain high positive SPLIT results for all four backbones. We observe AUC values above 0·90 for the backbones trained on ImageNet, all patients, and male patients, which further confirms that even high SPLIT responses are insufficient for drawing conclusions whether specific patient information is encoded in the backbone and used for making predictions about the presence of disease. SPLIT results on MIMIC-CXR are also provided in Figures 5 and 6 with more details in Tables S4 and S5. SPLIT results using a ResNet-34 on CheXpert are provided in the Supplementary Material in Figure S3 and Tables S6 and S7, all leading to similar findings.

**Performance of multitask model**

A multitask DenseNet-121 trained jointly for predicting race, sex, and presence of disease achieves a disease detection AUC of 0·86 (95% CI 0·86-0·87) and 0·85 (0·84-0·85) for 'no finding', and 0·86 (0·86-0·87) and 0·90 (0·89-0·90) for 'pleural effusion' on CheXpert and MIMIC-CXR, which is comparable to the performance of the corresponding single task models (cf. Tables 2 and 3). The multitask model also preserves similar high performance for both race and sex classification when compared to the corresponding single task models (cf. Figures 5 and 6). The fact that for the multitask model the performance for the individual prediction tasks is largely unaffected may suggest that in our setting race and sex information is not informative for detection of disease. Although, one may argue that if a (single task) disease detection model is already capable of implicity classifying sex and race, explicitly adding this information may actually not affect the performance. However, if such implicit encoding of race and sex in a single task model was similarly strong as the explicit use of the patient information in a multitask model, we would expect to find SPLIT responses on the single task disease detection backbone close to the performance of a multitask model when classifying race and sex. This is not the case, and in fact, there are large differences in average AUC between the multitask models and SPLIT of about 0·17 and 0·07 for race and sex classification, respectively. Here, the inspection of the feature representations discussed next aims to shed some light on the key differences in the way patient characteristics are encoded in the different models. Additional multitask results for a ResNet-34 on CheXpert with similar findings are provided in Table S1 in the Supplementary Material.

**Unsupervised feature exploration**

The inspection of the learned feature representations via PCA and t-SNE help to uncover the relationship between patient characteristics and disease detection, allowing us to assess whether a model maybe be under scenario B or C (cf. Figure 2). In Figure 7, we show a variety of different plots produced for the single task disease detection model trained on CheXpert applied to the resampled test-set for an unbiased assessment. The feature representations typically align well with the ground truth labels which can be observed clearly in the first mode of PCA separating samples that have different disease labels along the direction of largest variation. Similar is observed in the t-SNE embedding and the logit outputs for 'no finding' and 'pleural effusion'. This is because the features are learned to be discriminative with respect to the disease detection task. Samples with the same disease labels should obtain similar feature values and logit outputs, hence, visible grouping will emerge in the scatter plots. We also observe a separation in the marginal distributions along the dimension that best separates the data. The idea of the unsupervised exploration is then to inspect whether other types of information may show similar patterns which would indicate that the inspected information is related to the primary task. In such a case, we may have a strong indication that the model is under scenario C, and the inspected information may indeed be used for making predictions for the primary task. If no patterns emerge, neither in the embeddings of the feature representations nor in the logit outputs, and there are no obvious differences in the marginal distributions across subgroups, one may be carefully optimistic that the model is under scenario B. [25]

For the single task disease detection model in Figure 7, we observe no obvious patterns in the scatter plots for biological sex and race, neither in the feature embeddings nor in the logit outputs. This is despite the high SPLIT responses for race and sex classification (cf. Figures 5 and 6). In contrast, we observe some visual patterns for age where younger patients are grouped together with features and logit outputs aligning with 'no finding', which is not surprising as age strongly correlates with the presence of disease. We also observe a slight shift in the marginal distribution of 'pleural effusion' for Black patients (Figure 7d, fourth column) which explains the disparate performance in TPR/FPR.   Figure 8 provides the corresponding plots for the multitask model trained on CheXpert. Interesting observations can be made when comparing the feature representations of the multitask model with the ones obtained for the single task disease detection model, both trained and tested on the same set of patients. Recall that the multitask model is explicitly exposed to the patient characteristics during training. Here, we observe that biological sex and racial identity are strongly encoded in the feature representation of the multitask model, clearly separating the patients from different subgroups. However, we observe that the separation of subgroups is not aligned with the direction in feature space separating disease. Inspecting the PCA plots, in particular, we observe that biological sex becomes the predominant factor of variation encoded in the first PCA mode. Disease seems best separated in the second mode, while racial identity seems to be mostly encoded in the third and fourth mode of PCA. Given that the modes are orthogonal (which is a property of PCA), this may indicate that the most discriminative features for disease, sex, and race are largely unrelated on the output-level which would resemble scenario B in Figure 2. Similar to the single task model, no obvious patterns emerge for race and sex along the direction of disease. In the logit outputs of the disease prediction layer we again observe a shift for the marginal distribution of 'pleural effusion' for Black patients. More subtle interactions, however, may be missed by the visual inspection. For this reason, we also report p-values for statistical tests performed on all relevant pairs of subgroups when comparing their marginal distributions in PCA and logit space. The results are provided in Table 6. For the PCA modes that primarily encode disease (mode 1 for the single task model, mode 2 for the multitask model), the differences for all except one of the pairwise comparisons within race and sex subgroups are statistically not-significant. However, some tests indicate possible associations between disease and race in the feature space and the logit outputs. Asian patients show differences on the output for 'no finding' (when compared to White and Black patients). The tests also suggest an association of Black patients with the logit output of 'pleural effusion'. The statistical tests alone, however, are not sufficient indicators of disparate performance and need to be considered in combination with a comprehensive subgroup performance analysis.

## Discussion

The objective of this article was to highlight the general difficulties when trying to answer the question of what information is used when ML models make predictions. We have highlighted that SPLIT is insufficient and cannot provide definite answers. We argue that our proposed combination of test-set resampling, multitask learning, and unsupervised exploration of feature representations provides a comprehensive framework for assessing the relationship between the encoding of patient characteristics and disease detection. Our work fits well within the recent discussion of algorithmic auditing which specifically highlights subgroup analysis as an important component.[26,27] Here, our proposed framework offers methodology for gaining insights about the potential biases of AI models. We have demonstrated the utility of this framework with worked examples for both racial identity and biological sex in the context of image-based disease detection in chest X-ray. We found that previously reported disparities for 'no finding' disappeared when correcting for statistical subgroup differences using resampled test sets.[13] However, our analysis confirmed disparate performance for detecting 'pleural effusion' in Black patients. While we could not find strong evidence that race information is directly or indirectly used by the disease detection models, we have to remain careful as weak correlations between prediction tasks may not be detected due to limitations of using visual interpretation of the embedding plots.[25] The statistical analysis of the marginal distributions in PCA and logit space suggest some association between protected characteristics and prediction of disease. Identifying the underlying causes of disparate performance remains a challenge and will require further research. Besides the scarcity of data from underrepresented groups, which may limit the generalization capabilities of machine learning models, other sources of bias such as label noise are likely to contribute to subgroup disparities.[28,29] Label noise is of particular concern as it cannot be corrected for with strategies such as resampling, and instead, would require careful re-annotation of the dataset. A possible source of label noise is systematic misdiagnosis of certain subgroups causing a severe form of annotation bias. In the presence of multiple sources of bias, and the absence of specific knowledge (or assumptions) about the extent of bias, assessing model fairness is difficult.[30] It has been argued previously that integrating causal knowledge about the data generation process is key when studying performance disparities in machine learning algorithms.[17,31]

It is worth highlighting that there is a very active branch of machine learning research aiming to develop methodology that can prevent (or at least discourage) the use of protected characteristics for decision making.[32] Here, the goal is to learn fair representations that do not discriminate against groups or individuals.[33] A popular approach in 'fair ML' is adversarial training where a secondary task model is employed during training of the primary task.[34–36] The secondary, adversarial model acts as a critic to assess whether the learned feature representations contain features predictive of subgroups. This is related to SPLIT, with the difference that the secondary task directly affects the learning of the feature representations, similar to multitask learning, but encouraging the active removal of predictive features during training.[37] Other approaches focus on fair predictions by auditing and correcting performance disparities across subgroups during and even after the primary task model has been trained.[38,39] These advances in fair ML are encouraging, in particular, in cases where we can identify the causes of disparities (e.g., specific biases in the training data) and we have reliable information that the use of protected characteristics would be harmful. However, it is also worth highlighting that in many applications it may not be obvious that this is the case. In fact, some works in the fairness literature show that under certain circumstances in order to obtain fair machine learning models, the use of characteristics related to subgroup membership may be desired or even required.[40,41] Ethical limitations and regulatory requirements will also need to be consider for the development of such technical 'solutions'.[42] In any case, approaches for model inspection such as unsupervised exploration of feature representations, will remain important to establish whether certain information may or may not be used for making predictions.

In conclusion, we would like to re-emphasize the need for rigorous validation of AI including assessment of performance across vulnerable patient groups. Reporting guidelines such as CONSORT-AI[43], STARD-AI[44], and others, advocate for complete and transparent reporting when assessing AI performance. A detailed failure case analysis with results reported on relevant subgroups is essential for gaining trust and confidence in the use of AI for critical decision making. Disparities across patient groups can only be discovered with detailed performance analysis which requires access to representative and unbiased test sets.[45–47] We believe no machine learning training strategy or model inspection tool alone can ever replace the evidence gathered from well designed and executed validation studies and these will remain key in the context of safe and ethical use of AI.[1,48] New frameworks for auditing AI algorithms will likely play an important role for clinical deployment.[27,49,50] We would hope that our work makes a valuable contribution by complementing these frameworks, offering a practical and insightful approach for subgroup performance analysis of image-based disease detection models.

## Contributors

BG conceived and designed the study and conducted the experiments; BG, CJ, MB, and SW performed the statistical analysis, interpreted the results, and verified the underlying data; SW performed data pre-processing. The authors jointly conceptualized, edited, and reviewed the manuscript.

## Declaration of interests

BG is a part-time employee of HeartFlow and Kheiron Medical Technologies and holds stock options with both as part of the standard compensation package. He was a part-time employee at Microsoft Research until May, 2021 and a scientific advisor for Kheiron Medical Technologies until September, 2021. All other authors declare no competing interests.

**Data sharing**

All data used in this work is publicly available. The CheXpert imaging dataset together with the patient demographic information can be downloaded from https://stanfordmlgroup.github.io/competitions/chexpert/. The MIMIC-CXR imaging dataset can be downloaded from https://physionet.org/content/mimic-cxr-jpg/2.0.0/ with the corresponding patient demographic information available from https://physionet.org/content/mimiciv/1.0/.

All information to recreate the exact study sample used in this paper including splits of training, validation, and test sets, and all code that is required for replicating the results is available under an open source Apache 2.0 license in our dedicated GitHub repository https://github.com/biomedia-mira/chexploration.

**Acknowledgments**

We would like to thank the authors of the 'reading race' study for the discussions and for making their code and data openly available. We are also very grateful to Xiaoxuan Liu, Alastair Denniston, and Melissa McCradden for their valuable feedback and helpful discussions during the preparation of this manuscript. We also thank Anil Rao for suggesting the use of test-set resampling, Eike Petersen for related work, and the reviewers for their constructive comments. BG received funding from the European Research Council (ERC) under the European Union's Horizon 2020 research and innovation programme (Grant Agreement No. 757173, Project MIRA). SW is supported by the UKRI London Medical Imaging & Artificial Intelligence Centre for Value Based Healthcare. CJ is supported by Microsoft Research and EPSRC through Microsoft's PhD Scholarship Programme. MB is funded through an Imperial College London President's PhD Scholarship.

## Tables

Table 1. Characteristics of the study population

|  | CheXpert | | | | MIMIC-CXR | | | |
|---|---|---|---|---|---|---|---|---|
|  | All | White | Asian | Black | All | White | Asian | Black |
| Attribute | All data | | | | | | | |
| Patients | 42,884 | 33,338 | 6,642 | 2,904 | 43,209 | 32,756 | 1,881 | 8,572 |
| Scans | 127,118 | 99,027 (78) | 18,830 (15) | 9,261 (7) | 183,207 | 14,1865 (77) | 7,106 (4) | 34,236 (19) |
| Age (years) | 63 ± 17 | 64 ± 17 | 61 ± 17 | 56 ± 17 | 65 ± 17 | 66 ± 16 | 63 ± 18 | 61 ± 17 |
| Female | 52,436 (41) | 39,735 (40) | 8,132 (43) | 4,569 (49) | 85,193 (47) | 61,626 (43) | 31,22 (44) | 20,445 (60) |
| No finding | 10,916 (9) | 8,236 (8) | 1,716 (9) | 964 (10) | 56,615 (31) | 41,215 (29) | 22,21 (31) | 13,179 (38) |
| Pleural effusion | 51,574 (41) | 40,545 (41) | 7,953 (42) | 3,076 (33) | 46,224 (25) | 38,693 (27) | 19,16 (27) | 5,615 (16) |
|  | Training data | | | | | | | |
| Patients | 25,730 | 20,034 | 3,945 | 1,751 | 25,925 | 19,613 (76) | 1,110 (4) | 5,202 (20) |
| Scans | 76,205 | 59,238 (78) | 11,371 (15) | 5,596 (7) | 11,0280 | 86,098 (78) | 4,248 (4) | 19,934 (18) |
| Age (years) | 63 ± 17 | 64 ± 17 | 62 ± 17 | 56 ± 17 | 65 ± 17 | 66 ± 16 | 63 ± 18 | 60 ± 17 |
| Female | 31,432 (41) | 23,715 (40) | 4,976 (44) | 2,741 (49) | 51,138 (46) | 37,518 (44) | 1,897 (45) | 11,723 (59) |
| No finding | 6,514 (9) | 4,910 (8) | 1,046 (9) | 558 (10) | 34,530 (31) | 25,170 (29) | 1,330 (31) | 8,030 (40) |
| Pleural effusion | 31,015 (41) | 24,405 (41) | 4,754 (42) | 1,856 (33) | 27,806 (25) | 23,526 (27) | 11,08 (26) | 3,172 (16) |
|  | Validation data | | | | | | | |
| Patients | 4,288 | 3,348 | 666 | 274 | 4,321 | 3242 (75) | 209 (5) | 870 (20) |
| Scans | 12,673 | 9,945 (79) | 1,809 (14) | 919 (7) | 17,665 | 13,369 (76) | 776 (4) | 3,520 (20) |
| Age (years) | 62 ±17 | 63 ± 17 | 62 ± 17 | 55 ± 16 | 65 ± 17 | 67 ± 16 | 60 ± 22 | 62 ± 17 |
| Female | 5,030 (40) | 3,933 (40) | 667 (37) | 430 (47) | 8,245 (47) | 5,755 (43) | 336 (43) | 2,154 (61) |
| No finding | 1,086 (9) | 817 (8) | 175 (10) | 94 (10) | 5,393 (31) | 3,903 (29) | 232 (30) | 1,258 (36) |
| Pleural effusion | 5,049 (40) | 3,988 (40) | 738 (41) | 323 (35) | 4,575 (26) | 3,721 (28) | 230 (30) | 624 (18) |
|  | Test data | | | | | | | |
| Patients | 12,866 | 9,956 | 2,031 | 879 | 12,963 | 9,901 (76) | 562 (5) | 2,500 (19) |
| Scans | 38,240 | 29,844 (78) | 5,650 (15) | 2,746 (7) | 55,262 | 42,398 (77) | 2,082 (4) | 10,782 (19) |
| Age (years) | 63 ± 17 | 64 ± 17 | 61 ± 17 | 57 ± 16 | 65 ± 17 | 66 ± 16 | 65 ± 17 | 61 ± 17 |
| Female | 15,974 (42) | 12,087 (41) | 2,489 (44) | 1,348 (49) | 25,810 (47) | 18,353 (43) | 889 (43) | 6,568 (61) |
| No finding | 3,316 (9) | 2,509 (8) | 495 (9) | 312 (11) | 16,692 (30) | 12,142 (29) | 659 (32) | 3,891 (36) |
| Pleural effusion | 15,510 (41) | 12,152 (41) | 2,461 (44) | 897 (33) | 13,843 (25) | 11,446 (27) | 578 (28) | 1,819 (17) |

Breakdown of demographics over the set of patient scans by racial groups and training, validation and test splits. Percentages in brackets are with respect to the number of scans. We also report the number of unique patients for each group.

Table 2. Disease detection with DenseNet-121 trained and tested on CheXpert

| | **No finding** | | | | |
|---|---|---|---|---|---|
| | White | Asian | Black | Female | Male |
| Test-set | | | AUC (95% CI) | | |
| Original | 0·87 (0·86-0·88) | 0·88 (0·86-0·89) | 0·88 (0·87-0·90) | 0·87 (0·86-0·88) | 0·87 (0·86-0·88) |
| Resampled | 0·87 (0·86-0·88) | 0·87 (0·87-0·88) | 0·89 (0·88-0·89) | 0·87 (0·86-0·87) | 0·89 (0·88-0·89) |
| Multitask | 0·86 (0·86-0·87) | 0·86 (0·85-0·88) | 0·88 (0·86-0·90) | 0·86 (0·85-0·87) | 0·87 (0·86-0·88) |
| | | | TPR (95% CI) | | |
| Original | 0·79 (0·77-0·80) | 0·80 (0·76-0·83) | 0·84 (0·80-0·88) | 0·79 (0·77-0·82) | 0·79 (0·77-0·81) |
| Resampled | 0·80 (0·78-0·81) | 0·79 (0·78-0·81) | 0·81 (0·80-0·82) | 0·78 (0·76-0·79) | 0·82 (0·81-0·83) |
| Multitask | 0·78 (0·76-0·80) | 0·78 (0·74-0·82) | 0·82 (0·78-0·87) | 0·82 (0·80-0·84) | 0·76 (0·74-0·78) |
| | | | FPR (95% CI) | | |
| Original | 0·20 (0·20-0·20) | 0·20 (0·19-0·21) | 0·23 (0·21-0·24) | 0·20 (0·20-0·21) | 0·20 (0·20-0·20) |
| Resampled | 0·20 (0·20-0·21) | 0·20 (0·20-0·20) | 0·20 (0·19-0·20) | 0·20 (0·20-0·20) | 0·20 (0·20-0·20) |
| Multitask | 0·20 (0·20-0·20) | 0·19 (0·18-0·20) | 0·22 (0·21-0·24) | 0·23 (0·23-0·24) | 0·18 (0·17-0·18) |
| | | | Youden's J statistic (95% CI) | | |
| Original | 0·59 (0·57-0·60) | 0·60 (0·56-0·64) | 0·61 (0·57-0·65) | 0·59 (0·57-0·61) | 0·59 (0·57-0·61) |
| Resampled | 0·59 (0·58-0·61) | 0·59 (0·58-0·61) | 0·61 (0·60-0·63) | 0·58 (0·56-0·59) | 0·62 (0·61-0·63) |
| Multitask | 0·58 (0·56-0·59) | 0·59 (0·55-0·63) | 0·60 (0·56-0·65) | 0·58 (0·56-0·60) | 0·58 (0·56-0·60) |
| | **Pleural effusion** | | | | |
| | White | Asian | Black | Female | Male |
| Test-set | | | AUC (95% CI) | | |
| Original | 0·86 (0·86-0·87) | 0·88 (0·87-0·89) | 0·86 (0·85-0·88) | 0·87 (0·86-0·87) | 0·86 (0·86-0·87) |
| Resampled | 0·87 (0·86-0·87) | 0·88 (0·88-0·89) | 0·85 (0·84-0·85) | 0·87 (0·87-0·87) | 0·86 (0·86-0·86) |
| Multitask | 0·86 (0·86-0·87) | 0·88 (0·87-0·88) | 0·86 (0·85-0·88) | 0·87 (0·86-0·87) | 0·86 (0·86-0·87) |
| | | | TPR (95% CI) | | |
| Original | 0·77 (0·76-0·78) | 0·78 (0·76-0·80) | 0·71 (0·68-0·74) | 0·76 (0·75-0·78) | 0·77 (0·76-0·78) |
| Resampled | 0·78 (0·78-0·79) | 0·80 (0·80-0·81) | 0·72 (0·71-0·73) | 0·78 (0·77-0·79) | 0·76 (0·75-0·76) |
| Multitask | 0·77 (0·75-0·78) | 0·78 (0·77-0·80) | 0·69 (0·66-0·73) | 0·75 (0·73-0·76) | 0·78 (0·77-0·79) |
| | | | FPR (95% CI) | | |
| Original | 0·21 (0·20-0·21) | 0·19 (0·18-0·20) | 0·16 (0·14-0·17) | 0·20 (0·19-0·20) | 0·20 (0·20-0·21) |
| Resampled | 0·21 (0·21-0·21) | 0·21 (0·20-0·21) | 0·18 (0·18-0·19) | 0·20 (0·20-0·21) | 0·20 (0·19-0·20) |
| Multitask | 0·21 (0·20-0·21) | 0·20 (0·19-0·21) | 0·15 (0·14-0·17) | 0·18 (0·18-0·19) | 0·21 (0·21-0·22) |
| | | | Youden's J statistic (95% CI) | | |
| Original | 0·56 (0·55-0·57) | 0·59 (0·57-0·61) | 0·55 (0·52-0·59) | 0·57 (0·55-0·58) | 0·57 (0·55-0·58) |
| Resampled | 0·57 (0·56-0·58) | 0·59 (0·59-0·60) | 0·54 (0·52-0·55) | 0·57 (0·57-0·58) | 0·56 (0·55-0·57) |
| Multitask | 0·56 (0·55-0·57) | 0·59 (0·57-0·61) | 0·54 (0·51-0·58) | 0·56 (0·55-0·58) | 0·57 (0·55-0·58) |

Disease detection results reported separately for each race group and biological sex for 'no finding' (top) and 'pleural effusion' (bottom). TPR and FPR in subgroups are determined using a fixed decision threshold optimized over the whole patient population for a target FPR of 0·20.

Table 3. Disease detection with DenseNet-121 trained and tested on MIMIC-CXR

| | No finding | | | | |
|---|---|---|---|---|---|
| | White | Asian | Black | Female | Male |
| Test-set | AUC (95% CI) | | | | |
| Original | 0·85 (0·84-0·85) | 0·86 (0·84-0·88) | 0·85 (0·84-0·86) | 0·86 (0·85-0·86) | 0·84 (0·83-0·84) |
| Resampled | 0·84 (0·84-0·85) | 0·85 (0·84-0·85) | 0·86 (0·86-0·86) | 0·86 (0·85-0·86) | 0·84 (0·84-0·84) |
| Multitask | 0·84 (0·84-0·85) | 0·85 (0·84-0·87) | 0·85 (0·84-0·85) | 0·86 (0·85-0·86) | 0·84 (0·83-0·84) |
| | TPR (95% CI) | | | | |
| Original | 0·75 (0·74-0·75) | 0·74 (0·71-0·77) | 0·80 (0·79-0·82) | 0·78 (0·77-0·79) | 0·74 (0·72-0·74) |
| Resampled | 0·74 (0·74-0·75) | 0·73 (0·73-0·74) | 0·77 (0·76-0·78) | 0·77 (0·77-0·78) | 0·73 (0·72-0·73) |
| Multitask | 0·73 (0·72-0·74) | 0·76 (0·72-0·79) | 0·82 (0·81-0·83) | 0·77 (0·76-0·78) | 0·73 (0·72-0·74) |
| | FPR (95% CI) | | | | |
| Original | 0·19 (0·19-0·19) | 0·17 (0·15-0·19) | 0·25 (0·24-0·26) | 0·21 (0·21-0·21) | 0·19 (0·19-0·20) |
| Resampled | 0·20 (0·20-0·20) | 0·19 (0·18-0·19) | 0·21 (0·21-0·22) | 0·21 (0·21-0·21) | 0·19 (0·19-0·19) |
| Multitask | 0·18 (0·18-0·19) | 0·19 (0·17-0·21) | 0·27 (0·26-0·28) | 0·20 (0·20-0·21) | 0·20 (0·19-0·20) |
| | Youden's J statistic (95% CI) | | | | |
| Original | 0·55 (0·54-0·56) | 0·58 (0·54-0·61) | 0·55 (0·54-0·57) | 0·57 (0·56-0·58) | 0·54 (0·53-0·55) |
| Resampled | 0·54 (0·54-0·55) | 0·55 (0·54-0·55) | 0·56 (0·55-0·57) | 0·56 (0·55-0·57) | 0·54 (0·53-0·54) |
| Multitask | 0·55 (0·54-0·55) | 0·57 (0·53-0·60) | 0·55 (0·53-0·56) | 0·57 (0·56-0·58) | 0·54 (0·52-0·55) |
| | Pleural effusion | | | | |
| | White | Asian | Black | Female | Male |
| Test-set | AUC (95% CI) | | | | |
| Original | 0·89 (0·89-0·89) | 0·90 (0·88-0·91) | 0·91 (0·90-0·91) | 0·91 (0·90-0·91) | 0·89 (0·88-0·89) |
| Resampled | 0·89 (0·89-0·89) | 0·88 (0·88-0·89) | 0·90 (0·90-0·90) | 0·90 (0·89-0·90) | 0·88 (0·88-0·89) |
| Multitask | 0·89 (0·89-0·90) | 0·90 (0·88-0·91) | 0·91 (0·90-0·91) | 0·91 (0·90-0·91) | 0·89 (0·89-0·89) |
| | TPR (95% CI) | | | | |
| Original | 0·84 (0·84-0·85) | 0·84 (0·81-0·87) | 0·79 (0·77-0·81) | 0·83 (0·82-0·85) | 0·84 (0·83-0·85) |
| Resampled | 0·85 (0·85-0·86) | 0·82 (0·82-0·83) | 0·80 (0·79-0·80) | 0·83 (0·82-0·83) | 0·82 (0·82-0·83) |
| Multitask | 0·86 (0·85-0·87) | 0·81 (0·77-0·84) | 0·74 (0·71-0·76) | 0·86 (0·85-0·87) | 0·82 (0·81-0·83) |
| | FPR (95% CI) | | | | |
| Original | 0·22 (0·21-0·22) | 0·20 (0·18-0·22) | 0·15 (0·14-0·15) | 0·18 (0·18-0·18) | 0·22 (0·21-0·22) |
| Resampled | 0·22 (0·22-0·23) | 0·22 (0·21-0·22) | 0·16 (0·16-0·16) | 0·19 (0·19-0·19) | 0·21 (0·21-0·21) |
| Multitask | 0·23 (0·22-0·23) | 0·18 (0·16-0·20) | 0·11 (0·11-0·12) | 0·20 (0·20-0·20) | 0·20 (0·20-0·20) |
| | Youden's J statistic (95% CI) | | | | |
| Original | 0·63 (0·62-0·64) | 0·64 (0·60-0·67) | 0·64 (0·62-0·66) | 0·65 (0·64-0·67) | 0·62 (0·61-0·63) |
| Resampled | 0·63 (0·62-0·64) | 0·61 (0·60-0·62) | 0·64 (0·63-0·64) | 0·63 (0·63-0·64) | 0·62 (0·61-0·62) |
| Multitask | 0·63 (0·63-0·64) | 0·63 (0·59-0·66) | 0·62 (0·60-0·64) | 0·66 (0·65-0·67) | 0·62 (0·61-0·63) |

Disease detection results reported separately for each race group and biological sex for 'no finding' (top) and 'pleural effusion' (bottom). TPR and FPR in subgroups are determined using a fixed decision threshold optimized over the whole patient population for a target FPR of 0·20.

Table 4. Effect of training set on disease detection with Dense-121 tested on CheXpert

| | No finding | | | | |
|---|---|---|---|---|---|
| | White | Asian | Black | Female | Male |
| Train-set | AUC (95% CI) | | | | |
| CheXpert | 0·87 (0·86-0·88) | 0·88 (0·86-0·89) | 0·88 (0·87-0·90) | 0·87 (0·86-0·88) | 0·87 (0·86-0·88) |
| MIMIC-CXR | 0·85 (0·85-0·86) | 0·86 (0·84-0·88) | 0·88 (0·86-0·90) | 0·85 (0·84-0·86) | 0·86 (0·85-0·87) |
| | TPR (95% CI) | | | | |
| CheXpert | 0·79 (0·77-0·80) | 0·80 (0·76-0·83) | 0·84 (0·80-0·88) | 0·79 (0·77-0·82) | 0·79 (0·77-0·81) |
| MIMIC-CXR | 0·77 (0·76-0·79) | 0·78 (0·75-0·82) | 0·83 (0·79-0·87) | 0·78 (0·76-0·80) | 0·78 (0·76-0·80) |
| | FPR (95% CI) | | | | |
| CheXpert | 0·20 (0·20-0·20) | 0·20 (0·19-0·21) | 0·23 (0·21-0·24) | 0·20 (0·20-0·21) | 0·20 (0·20-0·20) |
| MIMIC-CXR | 0·20 (0·20-0·20) | 0·19 (0·18-0·20) | 0·23 (0·22-0·25) | 0·21 (0·20-0·21) | 0·19 (0·19-0·20) |
| | Youden's J statistic (95% CI) | | | | |
| CheXpert | 0·59 (0·57-0·60) | 0·60 (0·56-0·64) | 0·61 (0·57-0·65) | 0·59 (0·57-0·61) | 0·59 (0·57-0·61) |
| MIMIC-CXR | 0·57 (0·56-0·59) | 0·59 (0·55-0·63) | 0·60 (0·55-0·64) | 0·57 (0·55-0·59) | 0·59 (0·57-0·61) |
| | Pleural effusion | | | | |
| | White | Asian | Black | Female | Male |
| Train-set | AUC (95% CI) | | | | |
| CheXpert | 0·86 (0·86-0·87) | 0·88 (0·87-0·89) | 0·86 (0·85-0·88) | 0·87 (0·86-0·87) | 0·86 (0·86-0·87) |
| MIMIC-CXR | 0·85 (0·85-0·85) | 0·87 (0·86-0·88) | 0·85 (0·84-0·87) | 0·86 (0·85-0·86) | 0·85 (0·85-0·86) |
| | TPR (95% CI) | | | | |
| CheXpert | 0·77 (0·76-0·78) | 0·78 (0·76-0·80) | 0·71 (0·68-0·74) | 0·76 (0·75-0·78) | 0·77 (0·76-0·78) |
| MIMIC-CXR | 0·74 (0·73-0·75) | 0·76 (0·75-0·78) | 0·69 (0·66-0·72) | 0·75 (0·74-0·76) | 0·73 (0·72-0·74) |
| | FPR (95% CI) | | | | |
| CheXpert | 0·21 (0·20-0·21) | 0·19 (0·18-0·20) | 0·16 (0·14-0·17) | 0·20 (0·19-0·20) | 0·20 (0·20-0·21) |
| MIMIC-CXR | 0·20 (0·20-0·21) | 0·20 (0·19-0·22) | 0·17 (0·15-0·19) | 0·20 (0·20-0·21) | 0·20 (0·19-0·20) |
| | Youden's J statistic (95% CI) | | | | |
| CheXpert | 0·56 (0·55-0·57) | 0·59 (0·57-0·61) | 0·55 (0·52-0·59) | 0·57 (0·55-0·58) | 0·57 (0·55-0·58) |
| MIMIC-CXR | 0·54 (0·52-0·55) | 0·56 (0·54-0·58) | 0·52 (0·49-0·56) | 0·55 (0·53-0·56) | 0·53 (0·52-0·55) |

Disease detection results reported separately for each race group and biological sex for 'no finding' (top) and 'pleural effusion' (bottom). TPR and FPR in subgroups are determined using a fixed decision threshold optimized over the whole patient population for a target FPR of 0·20.

Table 5. Effect of training set on disease detection with Dense-121 tested on MIMIC-CXR

| | No finding | | | | |
|---|---|---|---|---|---|
| | White | Asian | Black | Female | Male |
| Train-set | AUC (95% CI) | | | | |
| MIMIC-CXR | 0·85 (0·84-0·85) | 0·86 (0·84-0·88) | 0·85 (0·84-0·86) | 0·86 (0·85-0·86) | 0·84 (0·83-0·84) |
| CheXpert | 0·82 (0·82-0·83) | 0·83 (0·81-0·85) | 0·83 (0·82-0·84) | 0·84 (0·83-0·84) | 0·81 (0·81-0·82) |
| | TPR (95% CI) | | | | |
| MIMIC-CXR | 0·75 (0·74-0·75) | 0·74 (0·71-0·77) | 0·80 (0·79-0·82) | 0·78 (0·77-0·79) | 0·74 (0·72-0·74) |
| CheXpert | 0·69 (0·68-0·70) | 0·73 (0·69-0·76) | 0·75 (0·74-0·77) | 0·74 (0·73-0·75) | 0·68 (0·67-0·69) |
| | FPR (95% CI) | | | | |
| MIMIC-CXR | 0·19 (0·19-0·19) | 0·17 (0·15-0·19) | 0·25 (0·24-0·26) | 0·21 (0·21-0·21) | 0·19 (0·19-0·20) |
| CheXpert | 0·19 (0·19-0·19) | 0·19 (0·17-0·21) | 0·25 (0·24-0·26) | 0·21 (0·21-0·21) | 0·19 (0·19-0·20) |
| | Youden's J statistic (95% CI) | | | | |
| MIMIC-CXR | 0·55 (0·54-0·56) | 0·58 (0·54-0·61) | 0·55 (0·54-0·57) | 0·57 (0·56-0·58) | 0·54 (0·53-0·55) |
| CheXpert | 0·50 (0·49-0·51) | 0·53 (0·49-0·57) | 0·51 (0·49-0·52) | 0·53 (0·52-0·54) | 0·49 (0·47-0·50) |
| | Pleural effusion | | | | |
| | White | Asian | Black | Female | Male |
| Train-set | AUC (95% CI) | | | | |
| MIMIC-CXR | 0·89 (0·89-0·89) | 0·90 (0·88-0·91) | 0·91 (0·90-0·91) | 0·91 (0·90-0·91) | 0·89 (0·88-0·89) |
| CheXpert | 0·88 (0·88-0·88) | 0·88 (0·87-0·90) | 0·90 (0·89-0·90) | 0·89 (0·89-0·90) | 0·88 (0·87-0·88) |
| | TPR (95% CI) | | | | |
| MIMIC-CXR | 0·84 (0·84-0·85) | 0·84 (0·81-0·87) | 0·79 (0·77-0·81) | 0·83 (0·82-0·85) | 0·84 (0·83-0·85) |
| CheXpert | 0·82 (0·82-0·83) | 0·82 (0·79-0·85) | 0·75 (0·73-0·77) | 0·82 (0·81-0·83) | 0·81 (0·80-0·82) |
| | FPR (95% CI) | | | | |
| MIMIC-CXR | 0·22 (0·21-0·22) | 0·20 (0·18-0·22) | 0·15 (0·14-0·15) | 0·18 (0·18-0·18) | 0·22 (0·21-0·22) |
| CheXpert | 0·22 (0·21-0·22) | 0·21 (0·19-0·23) | 0·14 (0·13-0·15) | 0·19 (0·18-0·19) | 0·21 (0·21-0·22) |
| | Youden's J statistic (95% CI) | | | | |
| MIMIC-CXR | 0·63 (0·62-0·64) | 0·64 (0·60-0·67) | 0·64 (0·62-0·66) | 0·65 (0·64-0·67) | 0·62 (0·61-0·63) |
| CheXpert | 0·61 (0·60-0·62) | 0·61 (0·57-0·64) | 0·61 (0·59-0·63) | 0·63 (0·62-0·64) | 0·60 (0·59-0·61) |

Disease detection results reported separately for each race group and biological sex for 'no finding' (top) and 'pleural effusion' (bottom). TPR and FPR in subgroups are determined using a fixed decision threshold optimized over the whole patient population for a target FPR of 0·20.

Table 6. Kolmogorov-Smirnov tests for marginal distributions on CheXpert

| | Single task disease detection model | | | | |
|---|---|---|---|---|---|
| | No find. / Pleur. eff. | White / Asian | Asian / Black | Black / White | Male / Female |
| Marginal | p-values | | | | |
| PCA mode 1 | <0·0001** | 1·00 | 0·019* | 0·42 | 1·00 |
| PCA mode 2 | 0·010* | 1·00 | <0·0001** | <0·0001** | 0·25 |
| PCA mode 3 | <0·0001** | 0·024* | 0·29 | 1·00 | 1·00 |
| PCA mode 4 | 0·12 | 1·00 | 0·42 | 0·17 | 0·15 |
| Logit 'no finding' | <0·0001** | 0·031* | 0·00098** | 1·00 | 0·18 |
| Logit 'pleural effusion' | <0·0001** | 1·00 | <0·0001** | 0·00025** | 0·24 |
| | Multitask disease detection model | | | | |
| | No find. / Pleur. eff. | White / Asian | Asian / Black | Black / White | Male / Female |
| Marginal | p-values | | | | |
| PCA mode 1 | <0·0001** | <0·0001** | 0·00025** | <0·0001** | <0·0001** |
| PCA mode 2 | <0·0001** | 0·18 | 0·20 | 1·00 | 0·38 |
| PCA mode 3 | 0·092 | <0·0001** | <0·0001** | <0·0001** | 0·029* |
| PCA mode 4 | 0·034* | <0·0001** | 0·077 | <0·0001** | 0·018* |
| Logit 'no finding' | <0·0001** | <0·0001** | <0·0001** | 0·70 | 0·00085** |
| Logit 'pleural effusion' | <0·0001** | 0·75 | 0·00017** | 0·018* | 0·034* |

Two-sample Kolmogorov-Smirnov tests are performed between the pairs of subgroups indicated in each column. The p-values are adjusted for multiple testing using the Benjamini-Yekutieli procedure and significance is determined at a 95% confidence level. Statistically significant results are marked with * p<0.05 and ** p<0.001.

# Figures

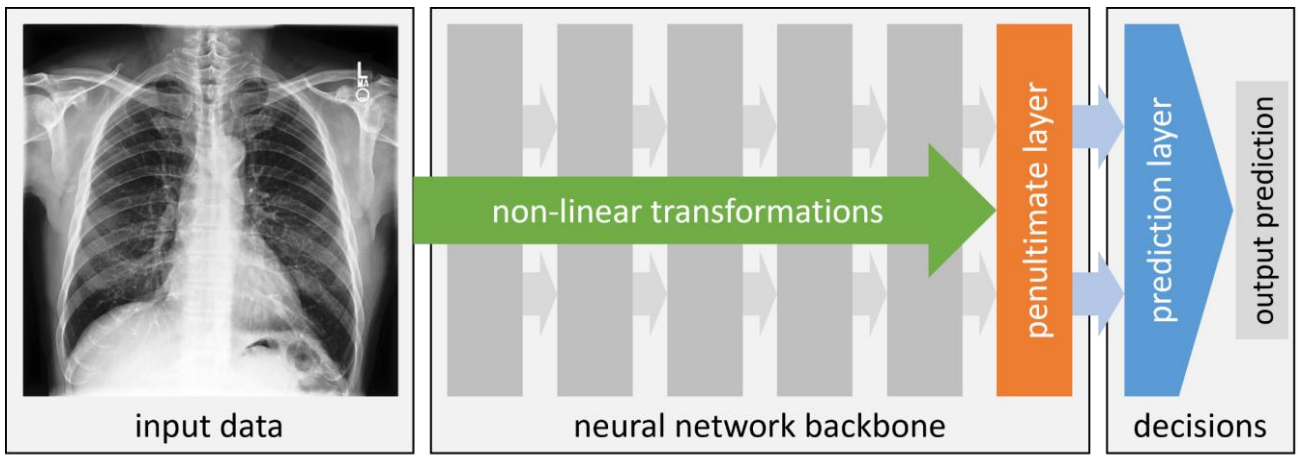

Figure 1. **Components of a deep neural network.**
An input medical scan is processed by a sequence of network layers in the so-called 'backbone' applying non-linear transformations whose parameters (or weights) are learned during training. This results in a complex feature representation at the penultimate layer which is then processed by the final prediction layer. The prediction layer assigns weights to each feature in the penultimate layer, aggregates the weighted features, and generates an output prediction. The prediction layer has the role of making the decision on what information is being used for making predictions.

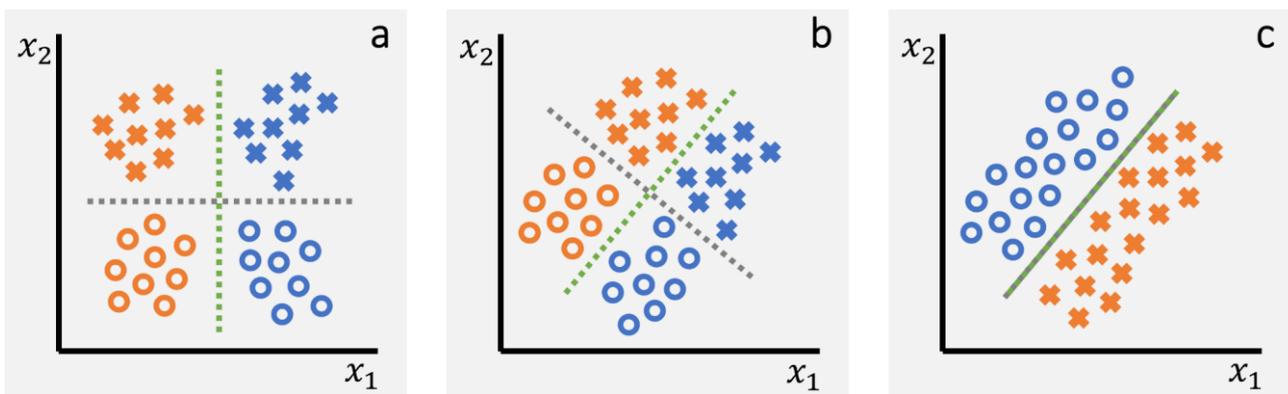

Figure 2. **Illustration of different inter-relationships of prediction tasks.**
**a** The two classification tasks of separating colors (blue vs orange) and shapes (crosses vs circles) are unrelated, both on the feature- and the output-level. The color classification can be performed by only considering feature $x_1$ while shape information is irrelevant. Similarly, shapes can be classified using feature $x_2$ with color being irrelevant. **b** The two tasks are related on a feature-level but not on their outputs. In both tasks, the features $x_1$ and $x_2$ need to be considered for classifying colors and shapes, however, shape information remains irrelevant for separating colors, and vice versa. While in both tasks the exact same features are being used, they are combined in different ways. **c** The two tasks are related both on a feature- and an output-level. Solving one of the tasks also solves the other. Shape and color information is highly correlated. The dashed green and gray lines indicate the optimal decision boundaries for color and shape classification, respectively.

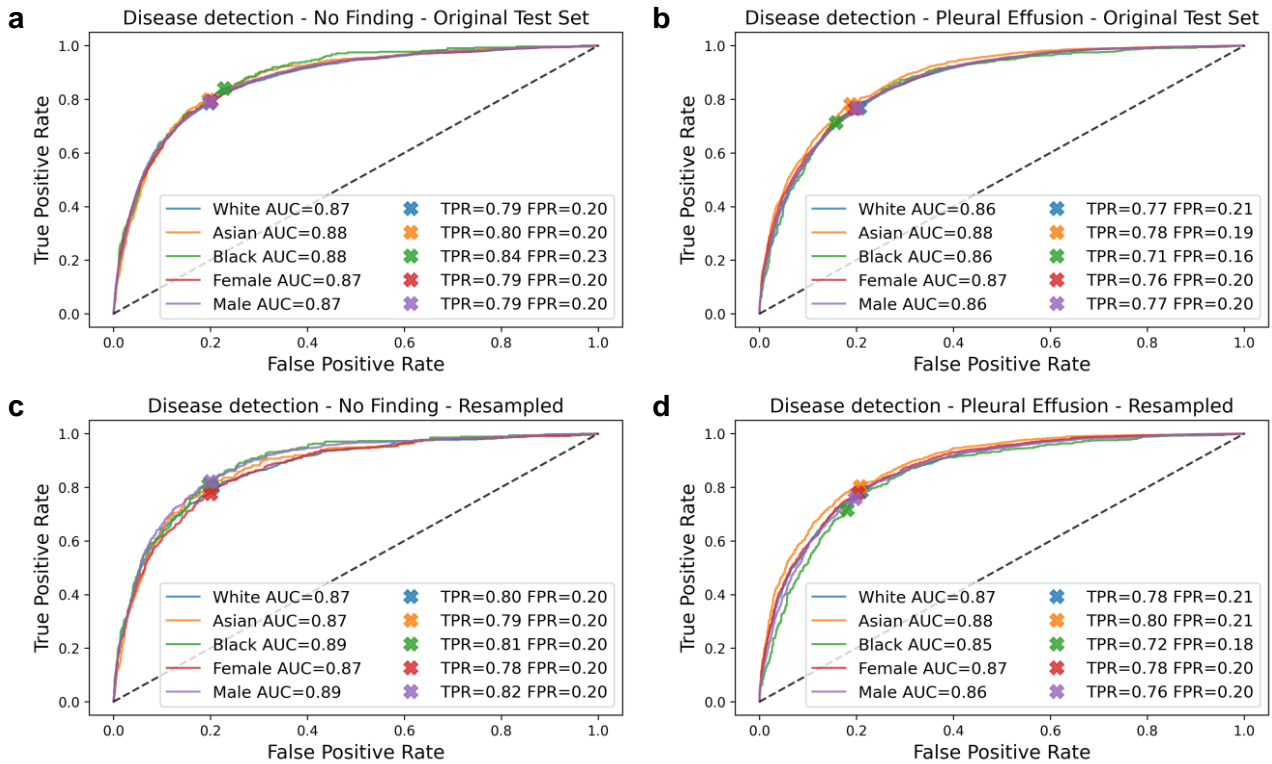

Figure 3. **Disease detection performance on original and resampled CheXpert test set.**
**a,b** ROC curves for the detection of 'no finding' and 'pleural effusion' with a DenseNet-121 trained on CheXpert and evaluated on the original test set. While AUC is largely consistent and ROC curves are similar in shape across subgroups, we observe a shift in TPR/FPR for Black patients compared to other groups. **c,d** ROC curves when evaluating the same model on a resampled test set correcting for race imbalance, age differences, and varying disease prevalence. The TPR/FPR shift disappears for 'no finding', while some smaller shifts remain for 'pleural effusion'.

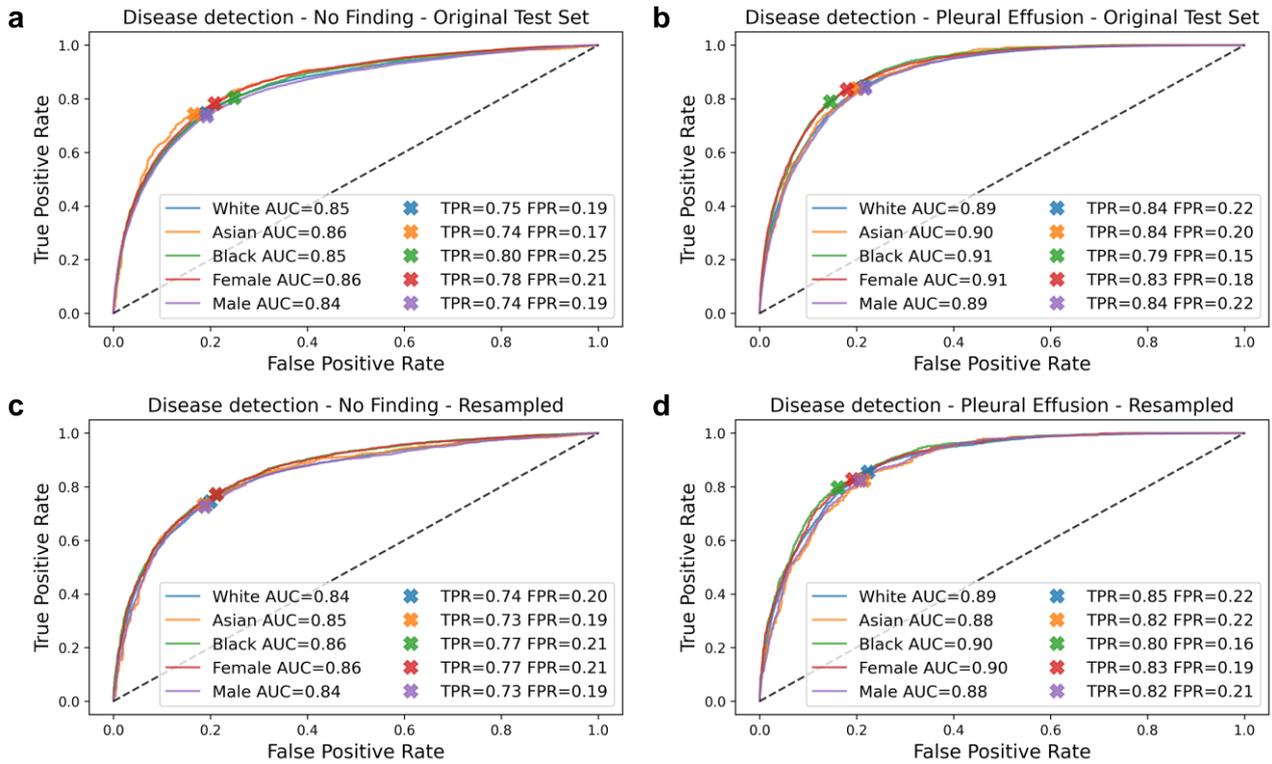

Figure 4. **Disease detection performance on original and resampled MIMIC-CXR test set.**
**a,b** ROC curves for the detection of 'no finding' and 'pleural effusion' with a DenseNet-121 trained on MIMIC-CXR and evaluated on the original test set. While AUC is largely consistent and ROC curves are similar in shape across subgroups, we observe a shift in TPR/FPR for Asian and Black patients for 'no finding', and multiple shifts for 'pleural effusion' compared to other groups. **c,d** ROC curves when evaluating the same model on a resampled test set correcting for race imbalance, age differences, and varying disease prevalence. The TPR/FPR shifts are largely reduced for 'no finding', while some shifts remain for 'pleural effusion'.

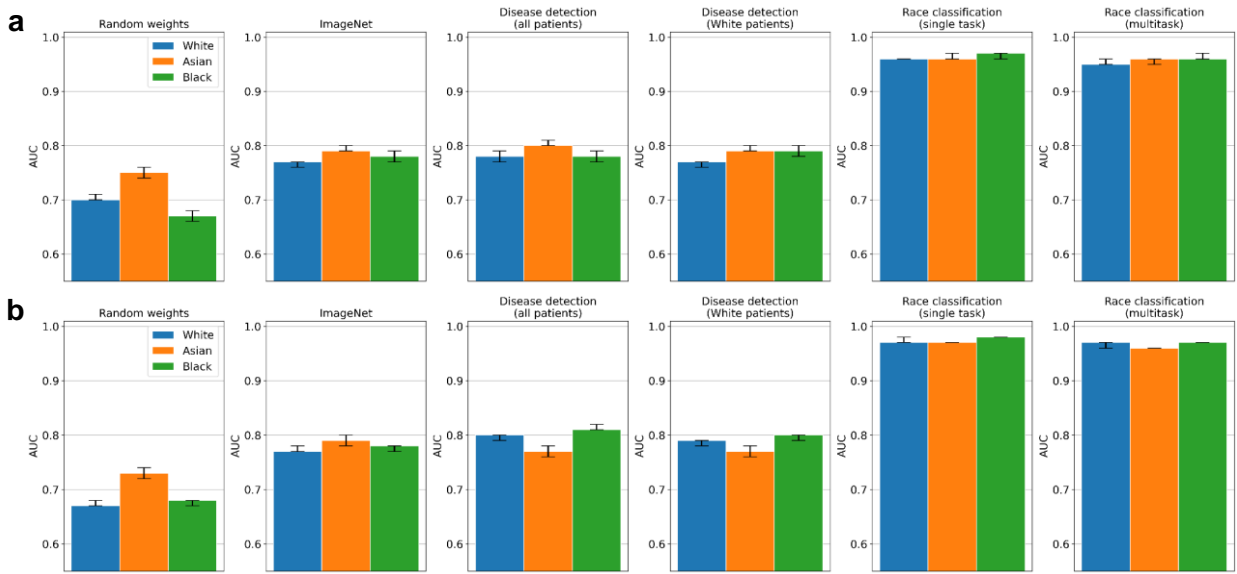

Figure 5. **Race classification with DenseNet-121 on CheXpert (a) and MIMIC-CXR (b)**
Classification performance is determined in a one-vs-rest approach for each racial group. The first four columns are the race classification results for SPLIT using different neural network backbones. Column five and six correspond to results from a single task race classification model and the multitask model trained jointly for disease, sex, and race classification. SPLIT performance on the ImageNet backbone and disease detection backbones is very similar.

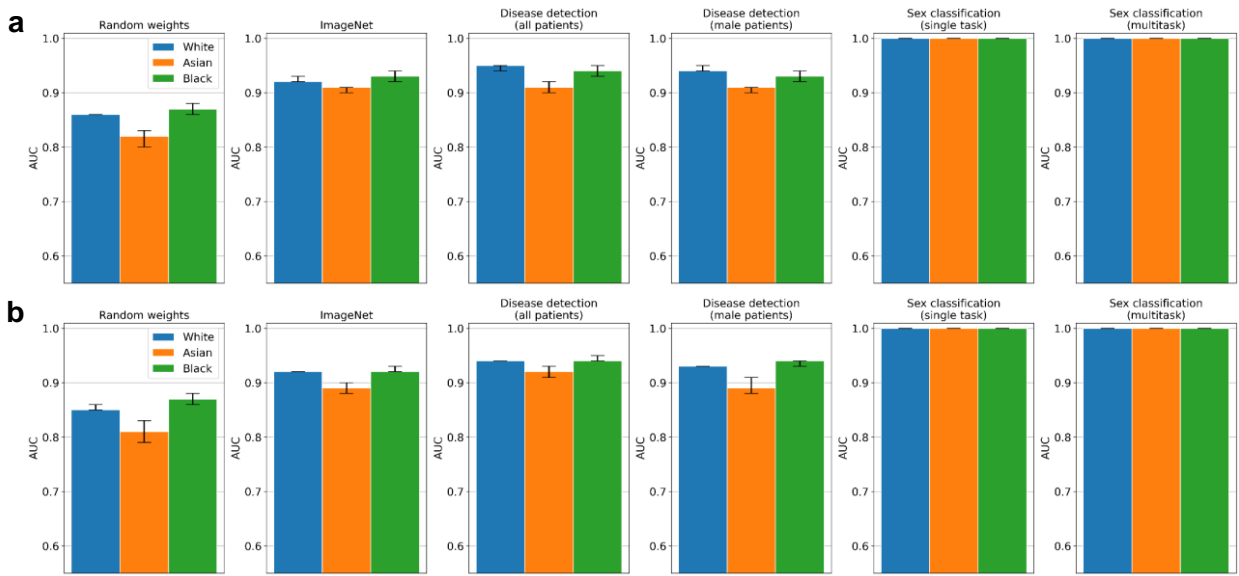

Figure 6. **Sex classification with DenseNet-121 on CheXpert (a) and MIMIC-CXR (b)**
The first four columns are the sex classification results for SPLIT using different neural network backbones. Column five and six correspond to results from a single task sex classification model and the multitask model trained jointly for disease, sex, and race classification. SPLIT performance on the ImageNet backbone and disease detection backbones is very similar.

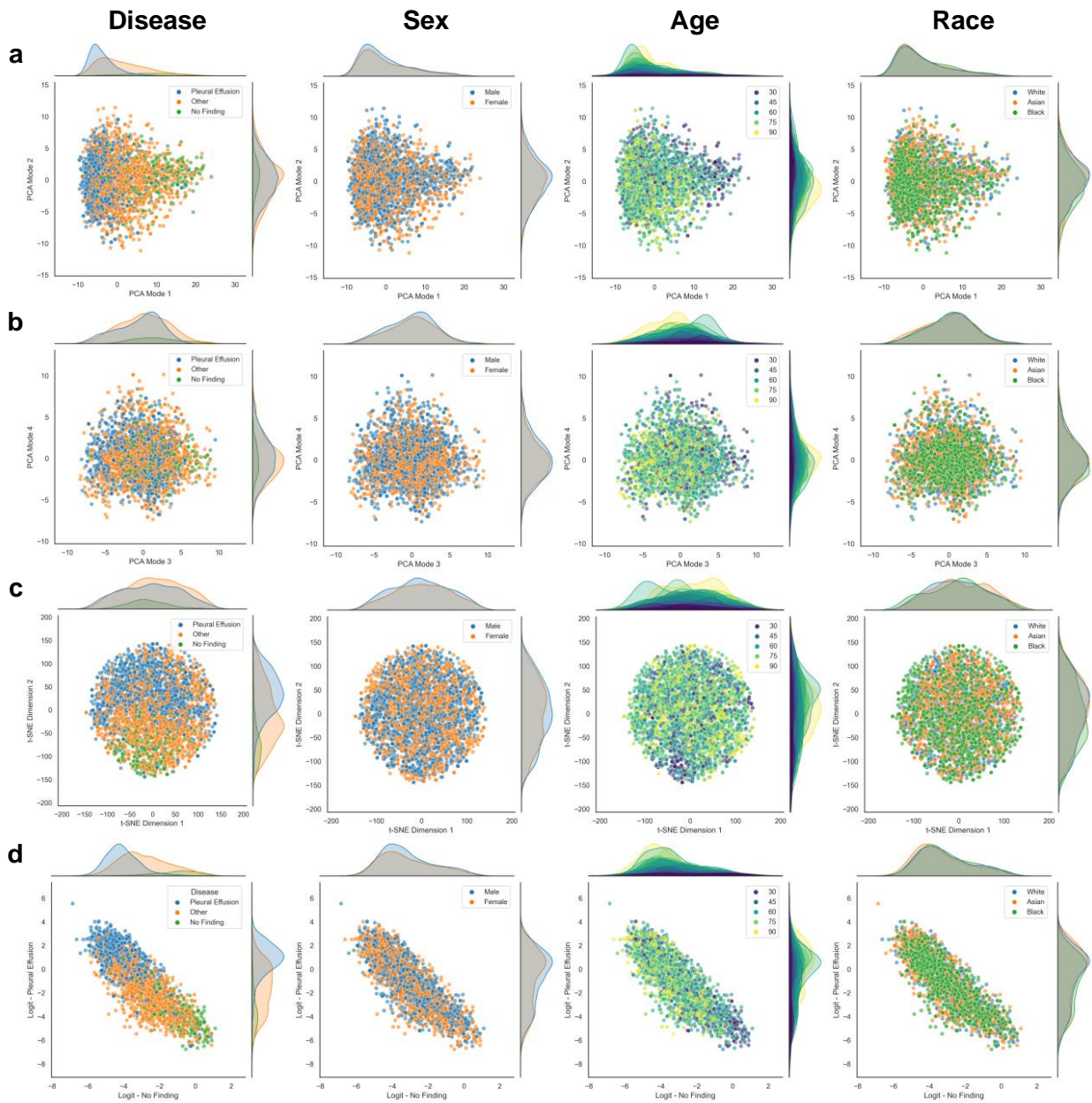

Figure 7. **Unsupervised exploration of feature representations (single task model).**
**a,b** Scatter plots with marginal distributions for the first four modes of PCA applied to feature representations of the CheXpert test data obtained with the single task disease detection model. A random set of 3,000 patients is shown with 1,000 samples from each racial group. Different types of information are overlaid in color from left to right including presence of disease, biological sex, age, and racial identity. **c** Scatter plots with marginal distributions for the t-SNE embedding. **d** Scatter plots with marginal distributions for the logit outputs produced by the model's prediction layer. No obvious patterns emerge for biological sex and race, while we observe a grouping of younger patients aligned with the 'no finding' label.

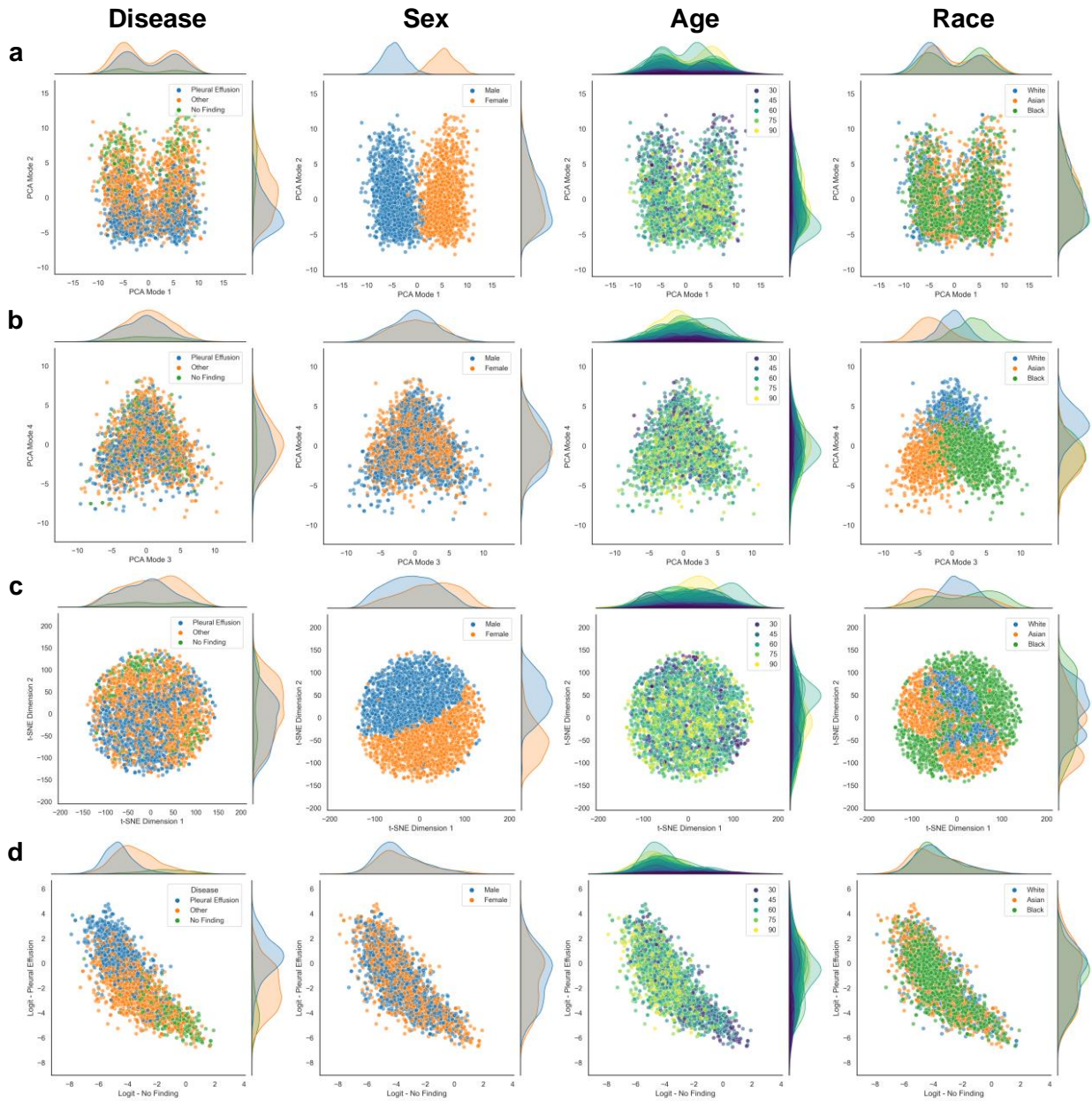

Figure 8. **Unsupervised exploration of feature representations (multitask model).**
**a,b** Scatter plots with marginal distributions for the first four modes of PCA applied to feature representations of the CheXpert test data obtained with the multitask disease detection model. A random set of 3,000 patients is shown with 1,000 samples from each racial group. Different types of information are overlaid in color from left to right including presence of disease, biological sex, age, and racial identity. **c** Scatter plots with marginal distributions for the t-SNE embedding. **d** Scatter plots with marginal distributions for the logit outputs produced by the model's prediction layer. The encoding of biological sex is clearly visible in the first mode of PCA, while disease features have the largest variation in the second mode. Racial identity seems to be primarily encoded in the third and fourth mode of PCA. Samples of same sex, race, and disease also form local groups in the t-SNE embedding. Despite the strong encoding of patient characteristics in the feature representations, no obvious patterns emerge for the logit outputs of the prediction layer, suggesting that race and sex are not relevant for the disease detection in this application.

# Supplementary Material

## Implementation details

For the deep neural network models, we use a PyTorch implementation of a DenseNet-121.[22] All chest X-ray images are resized using bi-linear interpolation to fit the required 224 x 224 pixel resolution. For the models trained for disease detection, race and sex classification, we initialize the network backbone with pre-trained ImageNet weights as provided in the torchvision module, except for the SPLIT models using backbones with random weights. We employ the Adam optimizer with default parameters and a learning rate of 0·001 for all our experiments.[51] We use the same configuration when training the prediction layers in the SPLIT experiments. We select the model checkpoint with the lowest cross-entropy loss on the validation set. To assess the variation in model performance due to randomness, we have trained the main disease detection models three times with different random seeds. The standard deviation across the reported performance metrics was 0·006 ± 0·005 on CheXpert and 0·004 ± 0·003 on MIMIC-CXR. We also explored another network architecture, ResNet-34[52], with the corresponding results reported in Tables S1, S6, S7, and Figure S3. Additional implementation details can be found in our code repository, including all trained models and scripts for test-set resampling and performance evaluation.

The input to PCA is a n-by-m feature matrix where n is the number of scans in the test set (38,240 for CheXpert and 55,262 for MIMIC-CXR), and m is the size of the output of the penultimate layer in the neural network (which is 1,024 for DenseNet-121 and 512 for ResNet-34). We use the PCA and t-SNE implementations provided in the Python-based machine learning library scikit-learn.[53] We initialize t-SNE with PCA embeddings as this has been shown to yield more consistent results.[25] We use PCA embeddings that preserve 99% of the variance of the feature representations. Otherwise, we use the default parameters for t-SNE. To improve the visibility in the scatter plots, we randomly sample 1,000 scans from each racial group, so in total 3,000 samples are shown in each scatter plot. PCA and t-SNE were computed on the entire test sets to capture the full variation in the subgroups. The random samples are used for visualization purposes only. For statistical testing, we use the two-sample Kolmogorov-Smirnov test as implemented in SciPy 1.8.0. P-values are adjusted for multiple testing using the Benjamini-Yekutieli procedure and significance is determined at a 95% confidence level.[54]

## Additional tables

Table S1. Disease detection with ResNet-34 trained and tested on CheXpert

| | White | Asian | Black | Female | Male |
|---|---|---|---|---|---|
| **No finding** | | | | | |
| Test-set | | | AUC (95% CI) | | |
| Original | 0·87 (0·86-0·88) | 0·87 (0·85-0·88) | 0·87 (0·85-0·89) | 0·86 (0·85-0·87) | 0·87 (0·86-0·88) |
| Resampled | 0·87 (0·86-0·87) | 0·86 (0·86-0·87) | 0·87 (0·87-0·88) | 0·85 (0·85-0·86) | 0·88 (0·88-0·89) |
| Multitask | 0·86 (0·85-0·87) | 0·86 (0·85-0·88) | 0·87 (0·85-0·89) | 0·86 (0·85-0·87) | 0·86 (0·86-0·87) |
| | | | TPR (95% CI) | | |
| Original | 0·79 (0·77-0·81) | 0·80 (0·77-0·84) | 0·79 (0·75-0·84) | 0·78 (0·75-0·80) | 0·81 (0·79-0·83) |
| Resampled | 0·79 (0·78-0·81) | 0·81 (0·79-0·82) | 0·77 (0·75-0·78) | 0·75 (0·74-0·77) | 0·82 (0·81-0·83) |
| Multitask | 0·76 (0·75-0·78) | 0·82 (0·79-0·86) | 0·84 (0·81-0·89) | 0·74 (0·72-0·77) | 0·81 (0·79-0·83) |
| | | | FPR (95% CI) | | |
| Original | 0·20 (0·20-0·20) | 0·19 (0·18-0·20) | 0·21 (0·20-0·23) | 0·20 (0·20-0·21) | 0·20 (0·19-0·20) |
| Resampled | 0·21 (0·20-0·21) | 0·20 (0·20-0·20) | 0·19 (0·19-0·20) | 0·20 (0·20-0·20) | 0·20 (0·20-0·20) |
| Multitask | 0·19 (0·19-0·19) | 0·24 (0·23-0·25) | 0·25 (0·23-0·26) | 0·17 (0·17-0·18) | 0·22 (0·21-0·22) |
| | | | Youden's J statistic (95% CI) | | |
| Original | 0·59 (0·57-0·61) | 0·61 (0·57-0·65) | 0·58 (0·53-0·63) | 0·57 (0·55-0·60) | 0·61 (0·59-0·63) |
| Resampled | 0·59 (0·57-0·60) | 0·61 (0·59-0·62) | 0·58 (0·56-0·59) | 0·56 (0·54-0·57) | 0·62 (0·61-0·63) |
| Multitask | 0·57 (0·56-0·59) | 0·59 (0·55-0·62) | 0·59 (0·56-0·64) | 0·57 (0·55-0·59) | 0·59 (0·57-0·61) |
| **Pleural effusion** | | | | | |
| | White | Asian | Black | Female | Male |
| Test-set | | | AUC (95% CI) | | |
| Original | 0·86 (0·86-0·86) | 0·87 (0·87-0·88) | 0·85 (0·84-0·87) | 0·86 (0·86-0·87) | 0·86 (0·86-0·86) |
| Resampled | 0·86 (0·86-0·86) | 0·88 (0·87-0·88) | 0·84 (0·83-0·84) | 0·86 (0·86-0·86) | 0·85 (0·85-0·86) |
| Multitask | 0·86 (0·85-0·86) | 0·88 (0·87-0·88) | 0·86 (0·84-0·87) | 0·86 (0·86-0·87) | 0·86 (0·86-0·86) |
| | | | TPR (95% CI) | | |
| Original | 0·76 (0·75-0·77) | 0·76 (0·74-0·78) | 0·68 (0·64-0·71) | 0·76 (0·75-0·77) | 0·76 (0·75-0·77) |
| Resampled | 0·77 (0·76-0·78) | 0·79 (0·78-0·79) | 0·68 (0·67-0·69) | 0·75 (0·74-0·76) | 0·74 (0·73-0·75) |
| Multitask | 0·76 (0·75-0·77) | 0·79 (0·77-0·81) | 0·68 (0·65-0·71) | 0·77 (0·76-0·78) | 0·75 (0·74-0·76) |
| | | | FPR (95% CI) | | |
| Original | 0·21 (0·20-0·21) | 0·19 (0·17-0·20) | 0·17 (0·15-0·19) | 0·20 (0·20-0·21) | 0·20 (0·19-0·20) |
| Resampled | 0·21 (0·20-0·21) | 0·20 (0·19-0·20) | 0·20 (0·19-0·20) | 0·21 (0·20-0·21) | 0·19 (0·19-0·20) |
| Multitask | 0·21 (0·20-0·21) | 0·20 (0·19-0·21) | 0·15 (0·13-0·16) | 0·21 (0·20-0·21) | 0·19 (0·19-0·20) |
| | | | Youden's J statistic (95% CI) | | |
| Original | 0·56 (0·55-0·57) | 0·58 (0·55-0·60) | 0·51 (0·47-0·54) | 0·56 (0·54-0·57) | 0·56 (0·55-0·57) |
| Resampled | 0·56 (0·55-0·57) | 0·59 (0·58-0·60) | 0·49 (0·47-0·49) | 0·54 (0·53-0·55) | 0·55 (0·54-0·55) |
| Multitask | 0·56 (0·55-0·56) | 0·59 (0·57-0·61) | 0·53 (0·49-0·56) | 0·56 (0·55-0·57) | 0·56 (0·55-0·57) |

Disease detection results reported separately for each race group and biological sex for 'no finding' (top) and 'pleural effusion' (bottom). TPR and FPR in subgroups are determined using a fixed decision threshold optimized over the whole patient population for a target FPR of 0·20.

Table S2. SPLIT for race classification with DenseNet-121 on CheXpert

| | White | Asian | Black |
|---|---|---|---|
| Neural network backbone | AUC (95% CI) | | |
| Random weights | 0·70 (0·70-0·71) | 0·75 (0·74-0·76) | 0·67 (0·66-0·68) |
| ImageNet | 0·77 (0·76-0·77) | 0·79 (0·79-0·80) | 0·78 (0·77-0·79) |
| Disease detection (all patients) | 0·78 (0·77-0·79) | 0·80 (0·80-0·81) | 0·78 (0·77-0·79) |
| Disease detection (White patients) | 0·77 (0·76-0·77) | 0·79 (0·79-0·80) | 0·79 (0·78-0·80) |
| Race classification (single task) | 0·96 (0·96-0·96) | 0·96 (0·96-0·97) | 0·97 (0·96-0·97) |
| Race classification (multitask) | 0·95 (0·95-0·96) | 0·96 (0·95-0·96) | 0·96 (0·96-0·97) |
| | TPR (95% CI) | | |
| Random weights | 0·64 (0·58-0·67) | 0·73 (0·66-0·78) | 0·64 (0·56-0·69) |
| ImageNet | 0·72 (0·65-0·73) | 0·76 (0·71-0·79) | 0·69 (0·66-0·73) |
| Disease detection (all patients) | 0·69 (0·67-0·72) | 0·73 (0·72-0·76) | 0·70 (0·62-0·73) |
| Disease detection (White patients) | 0·70 (0·65-0·75) | 0·71 (0·67-0·76) | 0·69 (0·67-0·77) |
| Race classification (single task) | 0·89 (0·89-0·91) | 0·90 (0·88-0·91) | 0·90 (0·89-0·91) |
| Race classification (multitask) | 0·89 (0·89-0·90) | 0·88 (0·87-0·90) | 0·90 (0·88-0·92) |
| | FPR (95% CI) | | |
| Random weights | 0·34 (0·28-0·37) | 0·35 (0·28-0·40) | 0·39 (0·31-0·43) |
| ImageNet | 0·32 (0·25-0·34) | 0·31 (0·26-0·34) | 0·28 (0·25-0·30) |
| Disease detection (all patients) | 0·27 (0·25-0·30) | 0·27 (0·26-0·30) | 0·28 (0·21-0·31) |
| Disease detection (White patients) | 0·30 (0·25-0·36) | 0·27 (0·23-0·32) | 0·26 (0·25-0·33) |
| Race classification (single task) | 0·10 (0·09-0·12) | 0·08 (0·08-0·10) | 0·08 (0·08-0·09) |
| Race classification (multitask) | 0·11 (0·11-0·13) | 0·09 (0·08-0·11) | 0·09 (0·08-0·11) |
| | Youden's J statistic (95% CI) | | |
| Random weights | 0·30 (0·29-0·31) | 0·38 (0·37-0·39) | 0·25 (0·24-0·27) |
| ImageNet | 0·40 (0·39-0·41) | 0·45 (0·44-0·46) | 0·41 (0·40-0·43) |
| Disease detection (all patients) | 0·42 (0·41-0·43) | 0·46 (0·45-0·47) | 0·42 (0·40-0·44) |
| Disease detection (White patients) | 0·40 (0·39-0·41) | 0·44 (0·43-0·46) | 0·43 (0·42-0·45) |
| Race classification (single task) | 0·79 (0·79-0·80) | 0·81 (0·80-0·82) | 0·82 (0·81-0·83) |
| Race classification (multitask) | 0·78 (0·77-0·78) | 0·79 (0·78-0·80) | 0·81 (0·80-0·82) |

Performance is determined in a one-vs-rest approach for each racial group. TPR and FPR in subgroups are determined using a fixed decision threshold optimized over the whole patient population for highest Youden's J statistic. For comparison, we also report the performance for single task and multitask neural networks trained specifically for classifying race, confirming the high accuracy reported in the literature.[8]

Table S3. SPLIT for sex classification with DenseNet-121 on CheXpert

| | White | Asian | Black |
|---|---|---|---|
| Neural network backbone | AUC (95% CI) | | |
| Random weights | 0·86 (0·86-0·86) | 0·82 (0·80-0·83) | 0·87 (0·86-0·88) |
| ImageNet | 0·92 (0·92-0·93) | 0·91 (0·90-0·91) | 0·93 (0·92-0·94) |
| Disease detection (all patients) | 0·95 (0·94-0·95) | 0·91 (0·90-0·92) | 0·94 (0·93-0·95) |
| Disease detection (male patients) | 0·94 (0·94-0·95) | 0·91 (0·90-0·91) | 0·93 (0·92-0·94) |
| Sex classification (single task) | 1·00 (1·00-1·00) | 1·00 (1·00-1·00) | 1·00 (1·00-1·00) |
| Sex classification (multitask) | 1·00 (1·00-1·00) | 1·00 (1·00-1·00) | 1·00 (1·00-1·00) |
| | TPR (95% CI) | | |
| Random weights | 0·79 (0·76-0·82) | 0·74 (0·70-0·83) | 0·85 (0·81-0·88) |
| ImageNet | 0·86 (0·84-0·88) | 0·85 (0·80-0·87) | 0·86 (0·83-0·89) |
| Disease detection (all patients) | 0·90 (0·87-0·90) | 0·80 (0·78-0·85) | 0·86 (0·82-0·90) |
| Disease detection (male patients) | 0·87 (0·86-0·89) | 0·85 (0·79-0·86) | 0·86 (0·82-0·89) |
| Sex classification (single task) | 0·98 (0·98-0·99) | 0·99 (0·98-0·99) | 0·98 (0·97-0·99) |
| Sex classification (multitask) | 0·98 (0·98-0·99) | 0·97 (0·97-0·99) | 0·99 (0·99-1·00) |
| | FPR (95% CI) | | |
| Random weights | 0·23 (0·20-0·26) | 0·25 (0·21-0·34) | 0·26 (0·21-0·29) |
| ImageNet | 0·17 (0·15-0·19) | 0·20 (0·15-0·22) | 0·16 (0·12-0·19) |
| Disease detection (all patients) | 0·14 (0·11-0·15) | 0·13 (0·12-0·19) | 0·12 (0·08-0·16) |
| Disease detection (male patients) | 0·13 (0·12-0·14) | 0·19 (0·13-0·20) | 0·15 (0·11-0·18) |
| Sex classification (single task) | 0·02 (0·01-0·02) | 0·03 (0·02-0·03) | 0·02 (0·01-0·04) |
| Sex classification (multitask) | 0·01 (0·01-0·02) | 0·02 (0·01-0·03) | 0·01 (0·00-0·02) |
| | Youden's J statistic (95% CI) | | |
| Random weights | 0·56 (0·55-0·57) | 0·49 (0·47-0·51) | 0·59 (0·57-0·62) |
| ImageNet | 0·69 (0·68-0·70) | 0·65 (0·64-0·68) | 0·70 (0·68-0·73) |
| Disease detection (all patients) | 0·76 (0·75-0·76) | 0·66 (0·65-0·69) | 0·74 (0·72-0·77) |
| Disease detection (male patients) | 0·75 (0·74-0·75) | 0·66 (0·64-0·68) | 0·71 (0·68-0·74) |
| Sex classification (single task) | 0·97 (0·96-0·97) | 0·96 (0·96-0·97) | 0·96 (0·95-0·97) |
| Sex classification (multitask) | 0·97 (0·97-0·97) | 0·95 (0·95-0·96) | 0·98 (0·98-0·99) |

Performance for classifying sex reported separately for each racial group. TPR and FPR in subgroups are determined using a fixed decision threshold optimized over the whole patient population for highest Youden's J statistic. For comparison, we also report the performance for single task and multitask neural networks trained specifically for sex classification, confirming high accuracy reported in the literature.[7]

Table S4. SPLIT for race classification with DenseNet-121 on MIMIC-CXR

| Neural network backbone | White | Asian | Black |
|---|---|---|---|
| | | AUC (95% CI) | |
| Random weights | 0·67 (0·67-0·68) | 0·73 (0·72-0·74) | 0·68 (0·67-0·68) |
| ImageNet | 0·77 (0·77-0·78) | 0·79 (0·78-0·80) | 0·78 (0·77-0·78) |
| Disease detection (all patients) | 0·80 (0·79-0·80) | 0·77 (0·76-0·78) | 0·81 (0·81-0·82) |
| Disease detection (White patients) | 0·79 (0·78-0·79) | 0·77 (0·76-0·78) | 0·80 (0·79-0·80) |
| Race classification (single task) | 0·97 (0·97-0·98) | 0·97 (0·97-0·97) | 0·98 (0·98-0·98) |
| Race classification (multitask) | 0·97 (0·96-0·97) | 0·96 (0·96-0·96) | 0·97 (0·97-0·97) |
| | | TPR (95% CI) | |
| Random weights | 0·66 (0·65-0·68) | 0·74 (0·66-0·78) | 0·59 (0·55-0·62) |
| ImageNet | 0·74 (0·68-0·76) | 0·76 (0·65-0·79) | 0·68 (0·66-0·72) |
| Disease detection (all patients) | 0·75 (0·72-0·79) | 0·71 (0·61-0·75) | 0·71 (0·69-0·73) |
| Disease detection (White patients) | 0·72 (0·69-0·75) | 0·69 (0·67-0·75) | 0·68 (0·67-0·73) |
| Race classification (single task) | 0·93 (0·91-0·94) | 0·89 (0·87-0·91) | 0·92 (0·91-0·93) |
| Race classification (multitask) | 0·92 (0·91-0·93) | 0·89 (0·87-0·91) | 0·90 (0·89-0·91) |
| | | FPR (95% CI) | |
| Random weights | 0·41 (0·39-0·43) | 0·40 (0·32-0·44) | 0·34 (0·30-0·36) |
| ImageNet | 0·33 (0·27-0·36) | 0·33 (0·22-0·35) | 0·27 (0·24-0·30) |
| Disease detection (all patients) | 0·31 (0·27-0·34) | 0·30 (0·21-0·35) | 0·24 (0·22-0·26) |
| Disease detection (White patients) | 0·29 (0·27-0·33) | 0·28 (0·26-0·34) | 0·24 (0·23-0·29) |
| Race classification (single task) | 0·10 (0·08-0·10) | 0·06 (0·06-0·08) | 0·07 (0·06-0·08) |
| Race classification (multitask) | 0·11 (0·10-0·13) | 0·11 (0·09-0·13) | 0·08 (0·07-0·09) |
| | | Youden's J statistic (95% CI) | |
| Random weights | 0·25 (0·25-0·26) | 0·34 (0·32-0·36) | 0·26 (0·25-0·27) |
| ImageNet | 0·41 (0·40-0·42) | 0·43 (0·41-0·45) | 0·42 (0·41-0·43) |
| Disease detection (all patients) | 0·45 (0·44-0·46) | 0·40 (0·39-0·43) | 0·47 (0·46-0·48) |
| Disease detection (White patients) | 0·42 (0·42-0·43) | 0·41 (0·39-0·43) | 0·45 (0·44-0·46) |
| Race classification (single task) | 0·83 (0·83-0·84) | 0·83 (0·81-0·84) | 0·85 (0·84-0·85) |
| Race classification (multitask) | 0·81 (0·80-0·81) | 0·78 (0·77-0·80) | 0·82 (0·82-0·83) |

Performance is determined in a one-vs-rest approach for each racial group. TPR and FPR in subgroups are determined using a fixed decision threshold optimized over the whole patient population for highest Youden's J statistic. For comparison, we also report the performance for single task and multitask neural networks trained specifically for classifying race, confirming the high accuracy reported in the literature.[8]

Table S5. SPLIT for sex classification with DenseNet-121 on MIMIC-CXR

|  | White | Asian | Black |
|---|---|---|---|
| Neural network backbone | AUC (95% CI) | | |
| Random weights | 0·85 (0·85-0·86) | 0·81 (0·79-0·83) | 0·87 (0·86-0·88) |
| ImageNet | 0·92 (0·92-0·92) | 0·89 (0·88-0·90) | 0·92 (0·92-0·93) |
| Disease detection (all patients) | 0·94 (0·94-0·94) | 0·92 (0·91-0·93) | 0·94 (0·94-0·95) |
| Disease detection (male patients) | 0·93 (0·93-0·93) | 0·89 (0·88-0·91) | 0·94 (0·93-0·94) |
| Sex classification (single task) | 1·00 (1·00-1·00) | 1·00 (1·00-1·00) | 1·00 (1·00-1·00) |
| Sex classification (multitask) | 1·00 (1·00-1·00) | 1·00 (1·00-1·00) | 1·00 (1·00-1·00) |
|  | TPR (95% CI) | | |
| Random weights | 0·77 (0·75-0·81) | 0·84 (0·78-0·87) | 0·77 (0·75-0·83) |
| ImageNet | 0·84 (0·83-0·86) | 0·79 (0·76-0·86) | 0·84 (0·82-0·87) |
| Disease detection (all patients) | 0·86 (0·85-0·89) | 0·84 (0·78-0·89) | 0·88 (0·85-0·90) |
| Disease detection (male patients) | 0·85 (0·83-0·87) | 0·82 (0·75-0·87) | 0·87 (0·84-0·89) |
| Sex classification (single task) | 0·98 (0·98-0·99) | 0·97 (0·96-0·98) | 0·98 (0·98-0·99) |
| Sex classification (multitask) | 0·98 (0·98-0·98) | 0·98 (0·96-0·99) | 0·98 (0·97-0·99) |
|  | FPR (95% CI) | | |
| Random weights | 0·23 (0·21-0·27) | 0·35 (0·28-0·37) | 0·20 (0·18-0·26) |
| ImageNet | 0·16 (0·15-0·17) | 0·17 (0·14-0·23) | 0·15 (0·12-0·18) |
| Disease detection (all patients) | 0·14 (0·13-0·17) | 0·15 (0·10-0·20) | 0·14 (0·11-0·16) |
| Disease detection (male patients) | 0·15 (0·13-0·17) | 0·19 (0·11-0·23) | 0·15 (0·12-0·17) |
| Sex classification (single task) | 0·01 (0·01-0·02) | 0·01 (0·01-0·03) | 0·01 (0·01-0·02) |
| Sex classification (multitask) | 0·02 (0·01-0·02) | 0·02 (0·01-0·03) | 0·02 (0·01-0·03) |
|  | Youden's J statistic (95% CI) | | |
| Random weights | 0·54 (0·53-0·55) | 0·49 (0·46-0·53) | 0·57 (0·56-0·59) |
| ImageNet | 0·68 (0·67-0·69) | 0·62 (0·59-0·66) | 0·69 (0·68-0·71) |
| Disease detection (all patients) | 0·72 (0·71-0·73) | 0·69 (0·66-0·72) | 0·74 (0·73-0·76) |
| Disease detection (male patients) | 0·70 (0·69-0·70) | 0·63 (0·61-0·67) | 0·72 (0·71-0·73) |
| Sex classification (single task) | 0·97 (0·97-0·97) | 0·96 (0·94-0·97) | 0·97 (0·96-0·97) |
| Sex classification (multitask) | 0·96 (0·96-0·96) | 0·96 (0·95-0·97) | 0·96 (0·95-0·96) |

Performance for classifying sex reported separately for each racial group. TPR and FPR in subgroups are determined using a fixed decision threshold optimized over the whole patient population for highest Youden's J statistic. For comparison, we also report the performance for single task and multitask neural networks trained specifically for sex classification, confirming high accuracy reported in the literature.[7]

Table S6. SPLIT for race classification with ResNet-34 on CheXpert

| | White | Asian | Black |
|---|---|---|---|
| Neural network backbone | AUC (95% CI) | | |
| Random weights | 0·60 (0·60-0·61) | 0·65 (0·64-0·65) | 0·58 (0·57-0·59) |
| ImageNet | 0·73 (0·73-0·74) | 0·77 (0·77-0·78) | 0·73 (0·72-0·74) |
| Disease detection (all patients) | 0·74 (0·73-0·74) | 0·76 (0·75-0·76) | 0·76 (0·75-0·77) |
| Disease detection (White patients) | 0·72 (0·72-0·73) | 0·75 (0·75-0·76) | 0·74 (0·72-0·75) |
| Race classification (single task) | 0·95 (0·95-0·96) | 0·96 (0·95-0·96) | 0·96 (0·96-0·97) |
| Race classification (multitask) | 0·95 (0·95-0·95) | 0·95 (0·95-0·96) | 0·96 (0·95-0·96) |
| | TPR (95% CI) | | |
| Random weights | 0·54 (0·46-0·62) | 0·68 (0·63-0·71) | 0·56 (0·35-0·60) |
| ImageNet | 0·63 (0·62-0·67) | 0·69 (0·68-0·76) | 0·71 (0·59-0·75) |
| Disease detection (all patients) | 0·66 (0·61-0·70) | 0·72 (0·66-0·80) | 0·75 (0·66-0·76) |
| Disease detection (White patients) | 0·65 (0·60-0·69) | 0·70 (0·67-0·78) | 0·63 (0·58-0·68) |
| Race classification (single task) | 0·89 (0·89-0·91) | 0·87 (0·86-0·89) | 0·89 (0·88-0·92) |
| Race classification (multitask) | 0·87 (0·86-0·89) | 0·87 (0·86-0·90) | 0·88 (0·86-0·90) |
| | FPR (95% CI) | | |
| Random weights | 0·39 (0·30-0·47) | 0·45 (0·41-0·49) | 0·44 (0·23-0·47) |
| ImageNet | 0·29 (0·28-0·33) | 0·28 (0·27-0·34) | 0·37 (0·24-0·40) |
| Disease detection (all patients) | 0·31 (0·25-0·34) | 0·33 (0·27-0·42) | 0·36 (0·28-0·37) |
| Disease detection (White patients) | 0·32 (0·27-0·36) | 0·32 (0·28-0·39) | 0·28 (0·23-0·32) |
| Race classification (single task) | 0·12 (0·12-0·14) | 0·08 (0·08-0·11) | 0·09 (0·08-0·12) |
| Race classification (multitask) | 0·12 (0·11-0·14) | 0·10 (0·09-0·13) | 0·09 (0·08-0·10) |
| | Youden's J statistic (95% CI) | | |
| Random weights | 0·16 (0·15-0·17) | 0·22 (0·21-0·24) | 0·12 (0·11-0·14) |
| ImageNet | 0·34 (0·33-0·36) | 0·41 (0·40-0·43) | 0·35 (0·33-0·37) |
| Disease detection (all patients) | 0·36 (0·35-0·37) | 0·38 (0·37-0·40) | 0·38 (0·37-0·40) |
| Disease detection (White patients) | 0·33 (0·32-0·35) | 0·38 (0·37-0·40) | 0·35 (0·33-0·37) |
| Race classification (single task) | 0·77 (0·76-0·78) | 0·78 (0·77-0·79) | 0·80 (0·79-0·81) |
| Race classification (multitask) | 0·75 (0·75-0·76) | 0·77 (0·76-0·78) | 0·79 (0·78-0·80) |

Performance is determined in a one-vs-rest approach for each racial group. TPR and FPR in subgroups are determined using a fixed decision threshold optimized over the whole patient population for highest Youden's J statistic. For comparison, we also report the performance for single task and multitask neural networks trained specifically for classifying race, confirming the high accuracy reported in the literature.[8]

Table S7. SPLIT for sex classification with ResNet-34 on CheXpert

| | White | Asian | Black |
|---|---|---|---|
| Neural network backbone | AUC (95% CI) | | |
| Random weights | 0·68 (0·67-0·69) | 0·65 (0·63-0·66) | 0·70 (0·68-0·72) |
| ImageNet | 0·91 (0·90-0·91) | 0·89 (0·88-0·90) | 0·91 (0·90-0·92) |
| Disease detection (all patients) | 0·91 (0·90-0·91) | 0·85 (0·84-0·86) | 0·91 (0·90-0·92) |
| Disease detection (male patients) | 0·91 (0·90-0·91) | 0·87 (0·86-0·88) | 0·90 (0·89-0·91) |
| Sex classification (single task) | 1·00 (1·00-1·00) | 1·00 (1·00-1·00) | 1·00 (1·00-1·00) |
| Sex classification (multitask) | 1·00 (1·00-1·00) | 1·00 (1·00-1·00) | 1·00 (1·00-1·00) |
| | TPR (95% CI) | | |
| Random weights | 0·64 (0·59-0·68) | 0·55 (0·48-0·76) | 0·53 (0·49-0·76) |
| ImageNet | 0·83 (0·81-0·87) | 0·81 (0·78-0·86) | 0·88 (0·81-0·90) |
| Disease detection (all patients) | 0·83 (0·81-0·84) | 0·74 (0·72-0·81) | 0·80 (0·77-0·88) |
| Disease detection (male patients) | 0·84 (0·82-0·85) | 0·82 (0·75-0·86) | 0·84 (0·78-0·86) |
| Sex classification (single task) | 0·99 (0·98-0·99) | 0·98 (0·97-0·99) | 0·98 (0·98-0·99) |
| Sex classification (multitask) | 0·98 (0·97-0·98) | 0·97 (0·96-0·98) | 0·98 (0·96-0·99) |
| | FPR (95% CI) | | |
| Random weights | 0·37 (0·32-0·41) | 0·33 (0·26-0·54) | 0·25 (0·20-0·47) |
| ImageNet | 0·17 (0·16-0·22) | 0·18 (0·15-0·23) | 0·21 (0·15-0·23) |
| Disease detection (all patients) | 0·17 (0·15-0·19) | 0·19 (0·17-0·26) | 0·13 (0·10-0·21) |
| Disease detection (male patients) | 0·19 (0·16-0·19) | 0·24 (0·17-0·28) | 0·20 (0·14-0·23) |
| Sex classification (single task) | 0·02 (0·01-0·02) | 0·03 (0·01-0·03) | 0·01 (0·01-0·02) |
| Sex classification (multitask) | 0·02 (0·02-0·02) | 0·03 (0·02-0·04) | 0·02 (0·01-0·04) |
| | Youden's J statistic (95% CI) | | |
| Random weights | 0·27 (0·26-0·28) | 0·22 (0·20-0·25) | 0·29 (0·26-0·32) |
| ImageNet | 0·66 (0·65-0·66) | 0·63 (0·61-0·65) | 0·67 (0·64-0·70) |
| Disease detection (all patients) | 0·66 (0·65-0·67) | 0·55 (0·53-0·57) | 0·67 (0·65-0·70) |
| Disease detection (male patients) | 0·66 (0·65-0·67) | 0·58 (0·56-0·60) | 0·64 (0·61-0·67) |
| Sex classification (single task) | 0·97 (0·97-0·97) | 0·96 (0·95-0·97) | 0·97 (0·96-0·98) |
| Sex classification (multitask) | 0·96 (0·96-0·96) | 0·94 (0·93-0·95) | 0·95 (0·94-0·97) |

Performance for classifying sex reported separately for each racial group. TPR and FPR in subgroups are determined using a fixed decision threshold optimized over the whole patient population for highest Youden's J statistic. For comparison, we also report the performance for single task and multitask neural networks trained specifically for sex classification, confirming high accuracy reported in the literature.[7]

## Additional figures

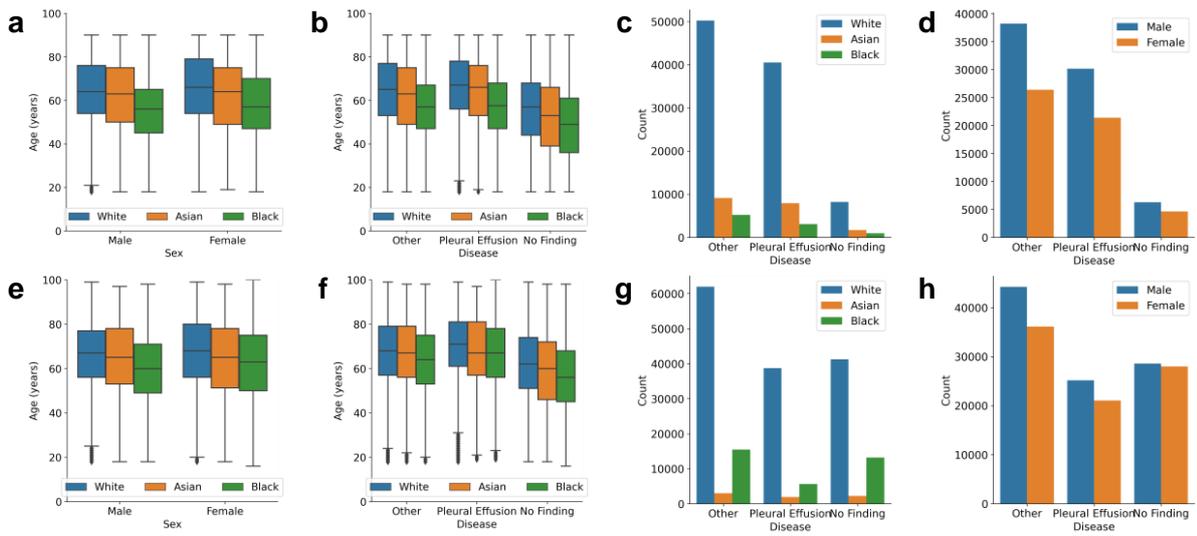

Figure S1. **Population characteristics of the original datasets.**
**a-d** CheXpert data shown at the top. **e-h** MIMIC-CXR shown at the bottom. **a,e** Age distribution over racial identity grouped by biological sex. **b,f** Age distribution over racial identity grouped by presence of disease. **c,g** Number of scans for each race grouped by presence of disease. **d,h** Number of scans for biological sex grouped by presence of disease. Whiskers in the box plot correspond to the largest and smallest samples inside the 1·5 interquartile range. Samples outside the boundary of the whiskers are plotted as outliers.

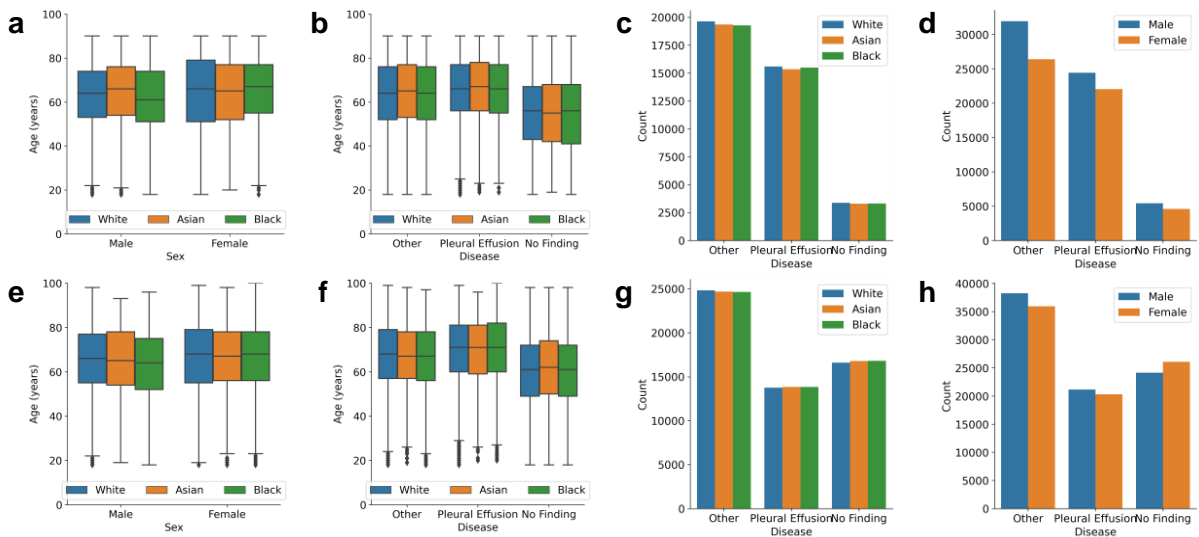

Figure S2. **Population characteristics of the resampled test-sets.**
**a-d** CheXpert data shown at the top. **e-h** MIMIC-CXR shown at the bottom. **a,e** Age distribution over racial identity grouped by biological sex. **b,f** Age distribution over racial identity grouped by presence of disease. **c,g** Number of scans for each race grouped by presence of disease. **d,h** Number of scans for biological sex grouped by presence of disease. Whiskers in the box plot correspond to the largest and smallest samples inside the 1·5 interquartile range. Samples outside the boundary of the whiskers are plotted as outliers.

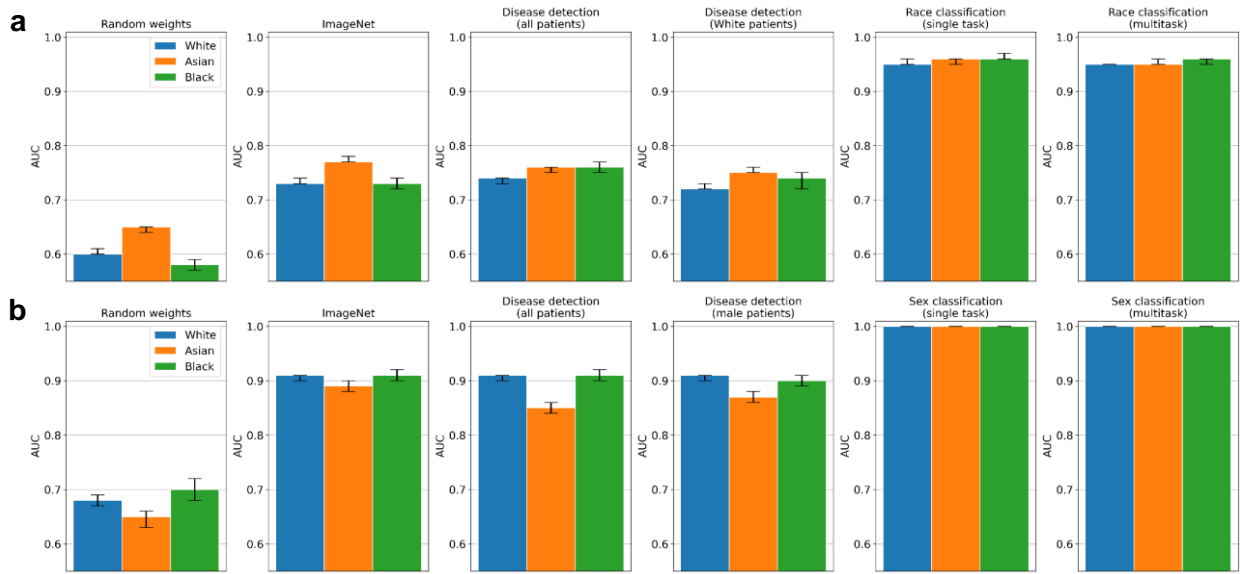

Figure S3. **Race (a) and sex (b) classification with ResNet-34 on CheXpert**
Race classification performance is determined in a one-vs-rest approach for each racial group. The first four columns are the race/sex classification results for SPLIT using different neural network backbones. Column five and six correspond to results from a single task race/sex classification model and the multitask model trained jointly for disease, sex, and race classification. SPLIT performance on the ImageNet backbone and disease detection backbones is very similar.